\documentclass[letterpaper]{article}
\usepackage{uai2018}
\usepackage[margin=1in]{geometry}
\usepackage{amsmath,amsfonts,amsthm,amssymb,xspace,bm}
\usepackage{verbatim,dsfont,mathtools,algorithm,algorithmic,url}
\usepackage{booktabs}

\usepackage{multirow}
\usepackage{xfrac}
\usepackage[numbers]{natbib}
\usepackage{wrapfig}
\usepackage{varwidth}
\usepackage{colortbl}
\usepackage{epstopdf}
\usepackage{appendix}
\usepackage{enumitem}
\usepackage{pifont}
\usepackage[table]{xcolor}
\usepackage{bbold}

\theoremstyle{plain}
\newtheorem{theorem}{Theorem}[section]
\newtheorem{lemma}[theorem]{Lemma}
\newtheorem*{lemma*}{Lemma}

\newtheorem{corollary}[theorem]{Corollary}

\newtheorem{definition}[theorem]{Definition}

\theoremstyle{definition}

\newtheorem{remark}{Remark}

\newcommand{\R}{\mathbb{R}}

\DeclareMathOperator{\trace}{\textsc{Tr}}

\DeclareMathOperator*{\argmax}{argmax}

\newcommand{\ip}[2]{\left\langle #1, #2 \right\rangle}

\newcommand{\norm}[1]{\left | #1\right |}
\newcommand{\dist}{{\rm{\textsc{Dist}}}}

\newcommand{\algo}{\texttt{BFGD}\xspace}

\newcommand{\X}{X}

\newcommand{\U}{U}

\newcommand{\f}{f}

\newcommand{\gradf}{\nabla \f}

\def\U{U}
\def\V{V}

\newcommand{\Uinit}{\U_0}
\newcommand{\Vinit}{\V_0}

\newcommand{\normf}[1]{\left |{#1}\right |_2}

\newcommand{\Xoptr}{X^{\star}_r}

\usepackage{times}

\title{Simple and practical algorithms for $\ell_p$-norm low-rank approximation}

\author{Anastasios Kyrillidis \\ IBM T.J. Watson Research Center \\ Rice University \\ \texttt{anastasios@rice.edu}} 

\begin{document}

\maketitle

\begin{abstract}
We propose practical algorithms for entrywise $\ell_p$-norm low-rank approximation, for $p = 1$ or $p = \infty$.
The proposed framework, which is non-convex and gradient-based, is easy to implement and typically attains
better approximations, faster, than state of the art.

From a theoretical standpoint, we show that the proposed scheme can attain $(1 + \varepsilon)$-OPT approximations.
Our algorithms are not hyperparameter-free: they achieve the desiderata only assuming algorithm's hyperparameters are known {\em apriori}---or are at least approximable.
\emph{I.e.}, our theory indicates what problem quantities need to be known, in order to get a good solution within polynomial time, and does not contradict to recent inapproximabilty results, as in \citep{song2017low}.
\end{abstract}

\section{Introduction}

We focus on the following optimization problem:
\begin{align} \label{eq:problem}
\min_{U \in \R^{m \times r}, V \in \R^{n \times r}} | M - UV^\top |_p,  \quad p \in \{1, \infty\}.
\end{align}
Here, $M \in \R^{m \times n}$ is a given input matrix of arbitrary rank, $r \leq \{m, n\}$ is the target rank, $(U, V)$ represent the variables such that $\texttt{rank}(UV^\top) \leq r$, and $| \cdot |_p$ denotes the $p$-th, \emph{entrywise}, matrix norm. 
In words, \eqref{eq:problem} is described as ``finding the factors of the best rank-$r$ approximation of $M$, with respect to the $\ell_p$-norm".
We denote such optimal factors $U^\star$ and $V^\star$, and their product $X^\star = U^\star V^{\star \top}$.
We focus on $p \in \{1, \infty\}$, since these instances are the most common found in practice, beyond the classic $p = 2$ (Frobenius) norm; we will use the terms ``Frobenius" and ``$\ell_2$" norm, interchangeably.

There are numerous applications where $\ell_1$- / $\ell_\infty$-norm low rank approximations are useful in practice.
First, the $\ell_1$-norm is more robust than the $\ell_2$-norm, and is suited in problem settings where Gaussian assumptions for noise models may not apply. 
$\ell_1$-norm low rank applications include robust PCA applications \citep{xu2010robust, candes2011robust, kyrillidis2012matrix, kyrillidis2014matrix, gu2016low, yi2016fast}, computer vision tasks such as background subtraction and motion detection \citep{turk1991eigenfaces, aanaes2002robust, meng2013cyclic}, detection of brain activation patterns \citep{qiu2014recursive}, and detection of anomalous behavior in dynamic networks \citep{qiu2014recursive}.
\footnote{
Closely related to the $\ell_1$-norm low-rank approximation is the problem of $\ell_1$-norm subspace recovery \citep{kwak2008principal}. Briefly,
it is well-known that, for $p = 2$ in \eqref{eq:problem}, the SVD solution is also the solution to the dual problem:
$U^\star = \argmax_{U \in \R^{m \times r}} | U^\top M |_2, ~\text{subject to}~ U^\top U = I$.
$V^\star$ is then set as $V^\star = U^{\star \top} M$; this can be easily proved due to the orthogonality of $U^\star$ \citep{golub2012matrix}.
Motivated by this dual formulation, $\ell_1$-norm subspace recovery is defined as 
\begin{align*}
U^\star = \argmax_{U \in \R^{m \times r}} | U^\top M |_1, ~~ \text{subject to}~~ U^\top U = I.
\end{align*} 
Algorithmic solutions to this criterion are usually greedy \citep{kwak2008principal}, even combinatorial  \citep{markopoulos2013some, markopoulos2014optimal}.
However, in this case, $U^\star$ does not necessarily resemble with that of \eqref{eq:problem} with $p = 1$ (up to orthogonal rotations).}


For the $\ell_\infty$-norm version of \eqref{eq:problem}, the problem cases are only a few.
\citep{poljak1993checking} considers the special case of $m = n$ and $r = \min\{m, n\} - 1$ as the problem of distance to robust non-singularity. 
\citep{goreinov2001maximal, goreinov2011quasioptimality} use the notion of $\ell_\infty$-norm low rank approximation for the maximal-volume concept in approximation, as well as for the skeleton approximation of a matrix.  
Finally, \citep{gillis2017low} identifies that \eqref{eq:problem} with $p = \infty$ can be used for the recovery of a low-rank matrix from a quantized $M$.

Despite the utility of \eqref{eq:problem}, its solution is not straightforward.
While \eqref{eq:problem} with $\ell_2$-norm has a closed-form solution via the Singular Value Decomposition (SVD), the same does not hold for $p \in \{1, \infty\}$.
Additionally, it has been proved that actually finding the exact solution to \eqref{eq:problem} can be exponentially complex:
\citep{gillis2015complexity} show that $\ell_1$-norm low rank matrix approximation is NP-hard, even for $r = 1$; 
further, under the exponential time hypothesis for \texttt{3SAT} problems, \citep{song2017low} provide a $\left(1 + \tfrac{1}{\log^{1+ \gamma} (\max\{m, n\})}\right)$-inapproximability result  for some hard instances $M$, where $\gamma > 0$ is an arbitrary small constant.
\citep{gillis2017low} proves the NP-completeness of \eqref{eq:problem} for $p = \infty$, using a reduction from \texttt{not-all-equal-3SAT}. 

The above restrict research to only approximations of \eqref{eq:problem}.
To the best of our knowledge only the works in \citep{chierichetti2017algorithms, song2017low} present polynomial and provably good approximation schemes:
\citep{song2017low} focuses mostly on the case of $\ell_1$-norm, and proves the existence of a $O(\log ( \min\{m, n\}) \cdot \texttt{poly}(r))$-approximation scheme with $O(\texttt{nnz}(M) + (m + n)\texttt{poly}(r))$ computational complexity.
\citep{chierichetti2017algorithms} extends the ideas in \citep{song2017low} for $\ell_p$-norms, where $p \in [1, \infty]$: 
there, the authors describe a $\texttt{poly}(r)$-approximation with $O\left(\texttt{poly}(m, n) (r \log \max\{m, n\})^r\right)$ computational complexity.
Both approaches are based on numerical linear algebra and sketching techniques.

Apart from the above provable schemes, there are numerous heuristics proposed for \eqref{eq:problem}, with no rigorous approximation guarantees.
Starting with $\ell_1$-norm, \citep{meng2013cyclic} propose a coordinate descent algorithm for \eqref{eq:problem}, where a sequence of alternating scalar minimization sub-problems are solved using a (weighted) median filter; see also \citep{kim2015efficient}.
Previously to that work, \citep{ke2003robust, ke2005robust} follow a similar approach, where each sub-problem is solved using linear or quadratic programming\footnote{In \citep{ke2003robust, ke2005robust}, there are some convergence guarantees for the alternating optimization scheme; however, there are no results w.r.t. whether we converge to a saddle point or local minimum, nor results on the convergence rate.}.
Inspired by \citep{wiberg1976computation}, \citep{eriksson2010efficient} propose a $\ell_1$-norm version of the Wiberg method; 
the resulting algorithm involves several matrix-matrix multiplications (even of size greater than the input matrix), and the solution of linear programming criteria, per iteration.
Cabral et al. use Augmented Lagrange Multipliers (ALM) method and handle the weighted $\ell_1$-norm low rank approximation problem in \citep{cabral2013unifying}; however, no non-asymptotic convergence guarantees are provided.
We note that most of the above heuristics are designed to handle \emph{missing data} in $M$ or the case of \emph{weighted} factorization; we plan to consider such cases for our future research directions.
For the $\ell_\infty$-norm case, we mention the recent work of Gillis et al. \citep{gillis2017low} that proposes a block coordinate descent method that operates in an alternating minimization fashion over subsets of variables in \eqref{eq:problem}.

\textbf{Our approach and main contributions:} 
Inspired by the recent advances on smooth non-convex optimization for matrix factorization \citep{sun2015guaranteed, zhao2015nonconvex, tu2015low, bhojanapalli2016dropping, park2016non, ge2016matrix, park2016provable, park2016finding, li2016recovery, li2016symmetry, tran2016extended, wang2017universal, ge2017no, kyrillidis2017provable}, we study the application of alternating gradient descent in \eqref{eq:problem}.
Despite its NP-hardness, this paper follows a more optimistic course and works towards deciphering the components/quantities that, if known a priori, could lead to a $(1 + \varepsilon)$-approximation for \eqref{eq:problem}.

Our approach is based on two techniques from optimization theory: 
$(i)$ the smoothing technique for non-smooth convex optimization by Nesterov \citep{nesterov2007smoothing, d2008smooth} (Section \ref{sec:func}), 
and $(ii)$ the recent theoretical results on finding the global minimum of matrix factorization problems using non-convex smooth methods (Section \ref{sec:BFGD}); see also references above.
Our theory relies on provably bounding the objective function in $\ell_1$- or $\ell_\infty$-norm by its smoothing counterpart (Sections \ref{sec:func}), using the provable performance of the non-convex algorithm (Section \ref{sec:BFGD}), and properly setting up the input parameters (Section \ref{sec:solver}).
Our guarantees assume that we can at least approximate the optimal function value of \eqref{eq:problem}, and that the optimal low-rank solution of the smoothed problem is well-conditioned; the latter assumption is required for a good initialization to be easily found. 
The above are summarized as: \vspace{-0.2cm}
\begin{itemize}[leftmargin=0.5cm]
\item Under assumptions, we provide a polynomial approximation algorithm for $p = \{1, \infty\}$ in \eqref{eq:problem} that achieves a $(1 + \epsilon)$-approximation guarantee. \vspace{-0.2cm}
\item We experimentally show that our scheme outperforms in practice state-of-the-art approaches. 
\end{itemize}
There are several questions that remain open and need further investigation. 
In Section \ref{sec:future}, we discuss what are the advantages and disadvantages of our approach and point to possible future research directions.

\section{Notation and assumptions}

\textit{Notation.} 
For matrices $X, Y \in \R^{m \times n}$, $\left \langle X, Y \right \rangle = \trace\left(X^\top Y \right)$ represents their inner product and $X \odot Y$ their Hadamard product.
We represent matrix norms as follows: $|X|_2 = \sqrt{\sum_{i = 1}^m \sum_{j = 1}^n |X_{ij}|^2}$ denotes the Frobenius (or $\ell_2$-) norm, $| X |_1 = \sum_{i = 1}^m \sum_{j = 1}^n |X_{ij}|$ denotes the entrywise $\ell_1$-norm, and $| X |_\infty = \max_{i,j} |X_{ij}|$ denotes the entrywise $\ell_\infty$-norm.
For the spectral norm, we use $\sigma_1(X)$; this also denotes the largest singular value of $X$.
For vectors, we use $\|x\|_2$ to denote its Euclidean $\ell_2$-norm.
For a differentiable function $f(X)$ with $X = UV^\top$, the gradient of $f$ w.r.t. $U$ and $V$ is $\gradf(X) V$ and $\gradf(X)^\top U$, respectively.  

\medskip
\noindent \textit{Assumptions.} 
For our discussion, we will need two well-known notions of convex analysis: (restricted) \emph{strong} convexity and (restricted) Lipschitz  gradient continuity. 

\begin{definition}{\label{prelim:def_00}}
Let $f: \R^{m \times n} \rightarrow \R$ be a convex differentiable function. 
Then, $f$ is (resp. restricted) gradient Lipschitz continuous with parameter $L$ if $\forall X, Y \in  \R^{m \times n}$ (resp. $\forall X, Y \in  \R^{m \times n}$ that are at most rank-$r$): 
\begin{equation}
f(Y) \leq f(X) + \ip{\gradf\left(X\right)}{Y - X} + \tfrac{L}{2} \norm{Y - X}_2^2.
\end{equation} 
\end{definition} 

\begin{definition}{\label{prelim:def_01}}
Let $f: \R^{m \times n} \rightarrow \R$ be convex and differentiable. Then, $f$ is (resp. restricted) $\mu$-strongly convex if $\forall X, Y \in  \R^{m \times n}$ (resp. $\forall X, Y \in  \R^{m \times n}$ that are at most rank-$r$): 
\begin{equation}\label{eq:sc}
f(Y) \geq f(X) + \ip{\gradf\left(X\right)}{Y - X} + \tfrac{\mu}{2} \norm{Y - X}_2^2.
\end{equation}
\end{definition}

\section{\algo for smooth objectives}{\label{sec:BFGD}}

Let us first succinctly describe the Bi-Factored Gradient Descent (\algo) algorithm \citep{park2016finding}, upon which our proposal is based. 
\algo is a non-convex gradient descent scheme for \emph{smooth} problems such as:
\begin{align}\label{eq:problem2}
\min_{U \in \R^{m \times r}, V \in \R^{n \times r}} f(UV^\top),
\end{align}
where $f$ is assumed to be convex, \emph{differentiable}, and at least have Lipschitz continuous gradients.
Observe that while $f$ is convex w.r.t. to any input $\in \R^{m \times n}$, motions over $U$ and $V$ jointly lead to non-convex optimization.
Such approaches have a long history and different variants have been proposed for  \eqref{eq:problem2}. 

For the rest of this section, we denote $X = UV^\top$ as the result of the factorization.
Also, let $\widehat{X}^\star$ be the optimal point of \eqref{eq:problem2}: if $\texttt{rank}(\widehat{X}^\star) = r$, then $\widehat{X}^\star = \widehat{X}^\star_r$; otherwise, denote its best rank-$r$ approximation (w.r.t. the $\ell_2$-norm) as $\widehat{X}^\star_r$.

\begin{algorithm}[!h]
	\caption{Bi-factored gradient descent (\algo)} \label{algo:FGD}
	\begin{small}
	\begin{algorithmic}[1]		
		\STATE \textbf{Input:} $r$, $T$, $\gamma~(\textit{e.g.}, \tfrac{1}{4})$, $C > 0 ~(\textit{e.g.}, C = 1)$, $\widehat{L}$. 
		\STATE Compute $X_0 := \sfrac{1}{\widehat{L}} \cdot \left(-\nabla f(0_{m \times n}) \right)$.
		\STATE Set $U_0 \in \R^{m \times r}, V_0 \in \R^{n \times r}$ s.t. $\X_0 = U_0V_0^\top$, via SVD.
		\FOR {$i=0$ to $T-1$}
			\STATE Set $\eta$ such that: $\eta \leq \tfrac{C}{15 \widehat{L} \norm{\begin{bmatrix} U_i~~ V_i \end{bmatrix}^\top }_2^2 + 3 \norm{\gradf(U_i V_i^\top) }_2}$. 
			\STATE \begin{itemize}
							\item If $f$ satisfies Definition \ref{prelim:def_00}: \textbf{Rule 1}
						\end{itemize}
														$$\footnotesize{\begin{bmatrix} U_{i+1} \\ V_{i+1} \end{bmatrix} = \begin{bmatrix} U_i \\ V_i \end{bmatrix} - \eta \begin{bmatrix} \gradf(U_i V_i^\top)\cdot V_i \\ \gradf(U_i V_i^\top)^\top  \cdot U_i \end{bmatrix}}$$ 
						\begin{itemize}
							\item If $f$ satisfies Definitions \ref{prelim:def_00}-\ref{prelim:def_01}: \textbf{Rule 2}
						\end{itemize}
														$$\!\!\!\!\!\!\!\scriptsize{\begin{bmatrix} U_{i+1} \\ V_{i+1} \end{bmatrix} = \begin{bmatrix} U_i \\ V_i \end{bmatrix} - \eta \begin{bmatrix} \gradf(U_i V_i^\top) V_i + \gamma U_i (U_i^\top U_i - V_i^\top V_i)  \\ \gradf(U_i V_i^\top)^\top  U_i - \gamma V_i (U_i^\top U_i - V_i^\top V_i)  \end{bmatrix}}$$ \vspace{-0.1cm}

		\ENDFOR
		\STATE \textbf{Output:} $\widehat{X} = U_T V_T^\top$. 	
	\end{algorithmic}
	\end{small}
\end{algorithm}

The pseudocode for \algo is provided in Algorithm \ref{algo:FGD} and obeys the following motions:
$(i)$ given a proper initialization $X_0 = U_0 V_0^\top$, 
and $(ii)$ a proper step size $\eta$,\footnote{In this work, we do not focus on the most efficient step size selections: \emph{e.g.}, the step size considered in this work varies per iteration, and it is less efficient than a constant step size selection as in \citep{bhojanapalli2016dropping, park2016finding}. However, in all cases, we could bound the varying step size with one that is constant.} 
\algo applies iteratively \textbf{Rule 1}
if $f$ satisfies only Definition \ref{prelim:def_00}, or \textbf{Rule 2} 
if $f$ also satisfies Definition \ref{prelim:def_01}.
The algorithm assumes an approximation of $L$---say $\widehat{L}$ and see \citep{bhojanapalli2016dropping}---and a good initialization point $(U_0, V_0)$.
For a more complete discussion of initialization $(U_0, V_0)$, we refer the reader to \cite{bhojanapalli2016dropping, park2016finding}; we briefly discuss this issue in Section \ref{sec:solver}.

An important issue in optimizing $f$ over $(U, V)$ is the existence of non-unique possible factorizations for a given $X$. 
We need a notion of distance to the low-rank solution $\widehat{X}^\star_r$ over the factors. 
Similar to \citep{tu2015low, park2016finding}, we focus on the set of ``equally-footed'' factorizations:
\begin{small}
\begin{align}
\widehat{\mathcal{X}}^\star_r = \Big\{ &\left(\widehat{U}^\star, \widehat{V}^\star\right):\widehat{U}^\star \in \R^{m \times r},\widehat{V}^\star \in \R^{n \times r},\widehat{U}^\star \widehat{V}^{\star ^\top} = \widehat{X}^\star_r, \nonumber \\ &\sigma_i(\widehat{U}^\star) = \sigma_i(\widehat{V}^\star) = \sigma_i( \widehat{X}^\star_r)^{1/2}, \forall i \in [r]\Big\}. \label{prelim:eq_footing}
\end{align}
\end{small}
Given a pair $(U,V)$, we define the distance to $\widehat{X}^\star_r$ as:
\begin{align*}
\dist\left(U,V; \widehat{X}^\star_r\right) = \min_{(\widehat{U}^\star, \widehat{V}^\star) \in \widehat{\mathcal{X}}^\star_r} \normf{\begin{bmatrix} U \\ V \end{bmatrix} - \begin{bmatrix} \widehat{U}^\star \\ \widehat{V}^\star \end{bmatrix}}.
\end{align*}

Algorithm \ref{algo:FGD} has local convergence guarantees, when $f$ is $\mu$-strongly convex and has $L$-Lipschitz continuous gradients, according to the following theorem:\footnote{In this work, we will borrow only the sublinear rate results in \citep{park2016finding}, since that result alone is sufficient to lead to polynomial algorithms for \eqref{eq:problem}. Using the linear convergence rate result in \citep{park2016finding} is left for the extension of this work.}
\begin{theorem}[Theorem 4.4 in \citep{park2016finding}] \label{thm:smooth}
Let $\kappa = L/\mu$.
If the initial point $\X_0 = \U_0\V_0^\top$, satisfies $\dist(\Uinit, \Vinit ; \Xoptr) \leq \tfrac{\sqrt{2} \cdot \sigma_r(\Xoptr)^{1/2}}{10}$, then \algo converges with rate $O(1/T)$:
\begin{align*}
f(U_T V_T^\top) - f(\widehat{U}^\star \widehat{V}^{\star \top}) \le \tfrac{10 \cdot \dist(\Uinit,\Vinit;\widehat{X}^\star_r)^2}{\eta T}
\end{align*}
\end{theorem}


\section{Charbonnier approximation and the \texttt{logsumexp} function}{\label{sec:func}
\begin{figure*}[!ht]
  \centering
    	\includegraphics[width=0.32\textwidth]{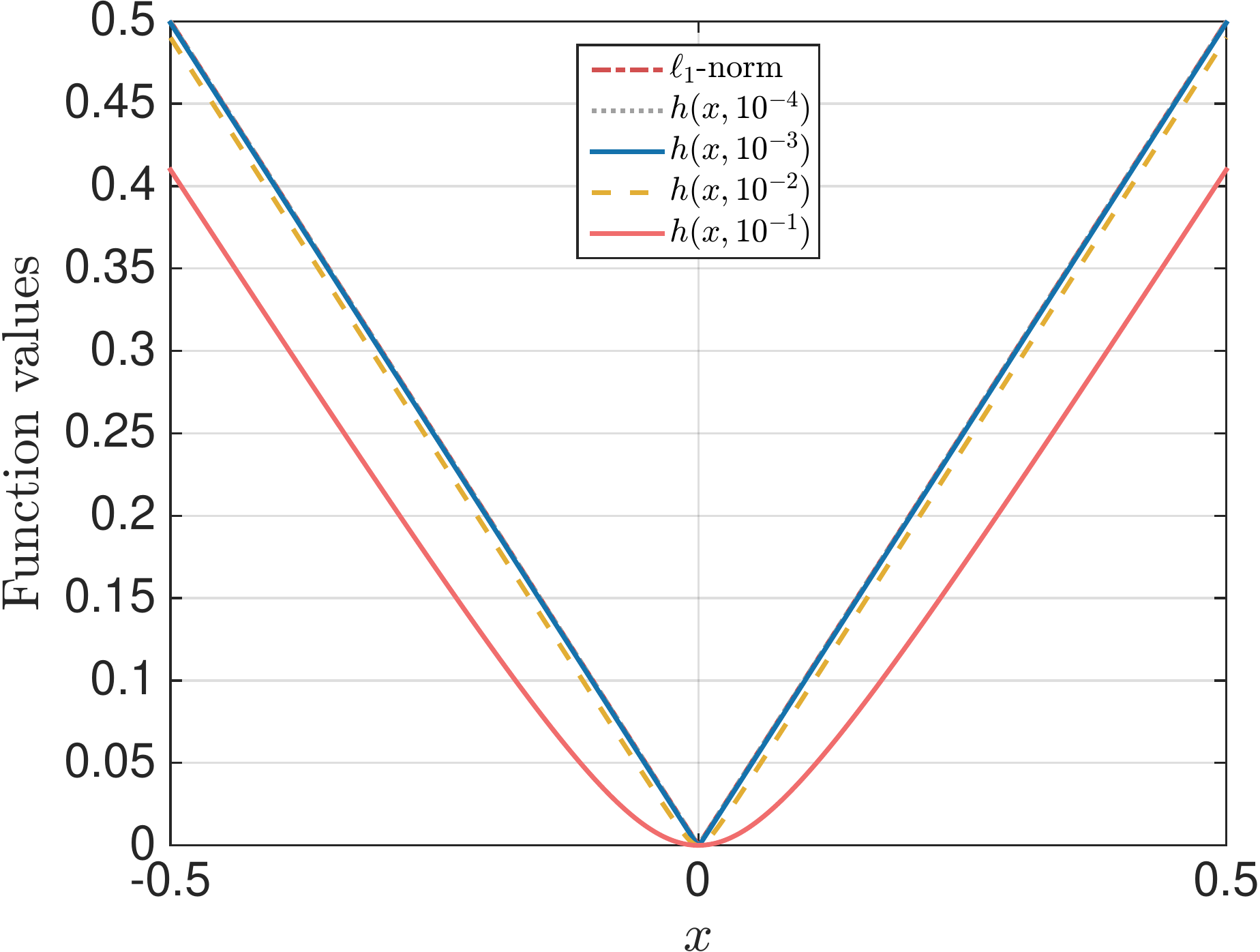}
	\includegraphics[width=0.27\textwidth]{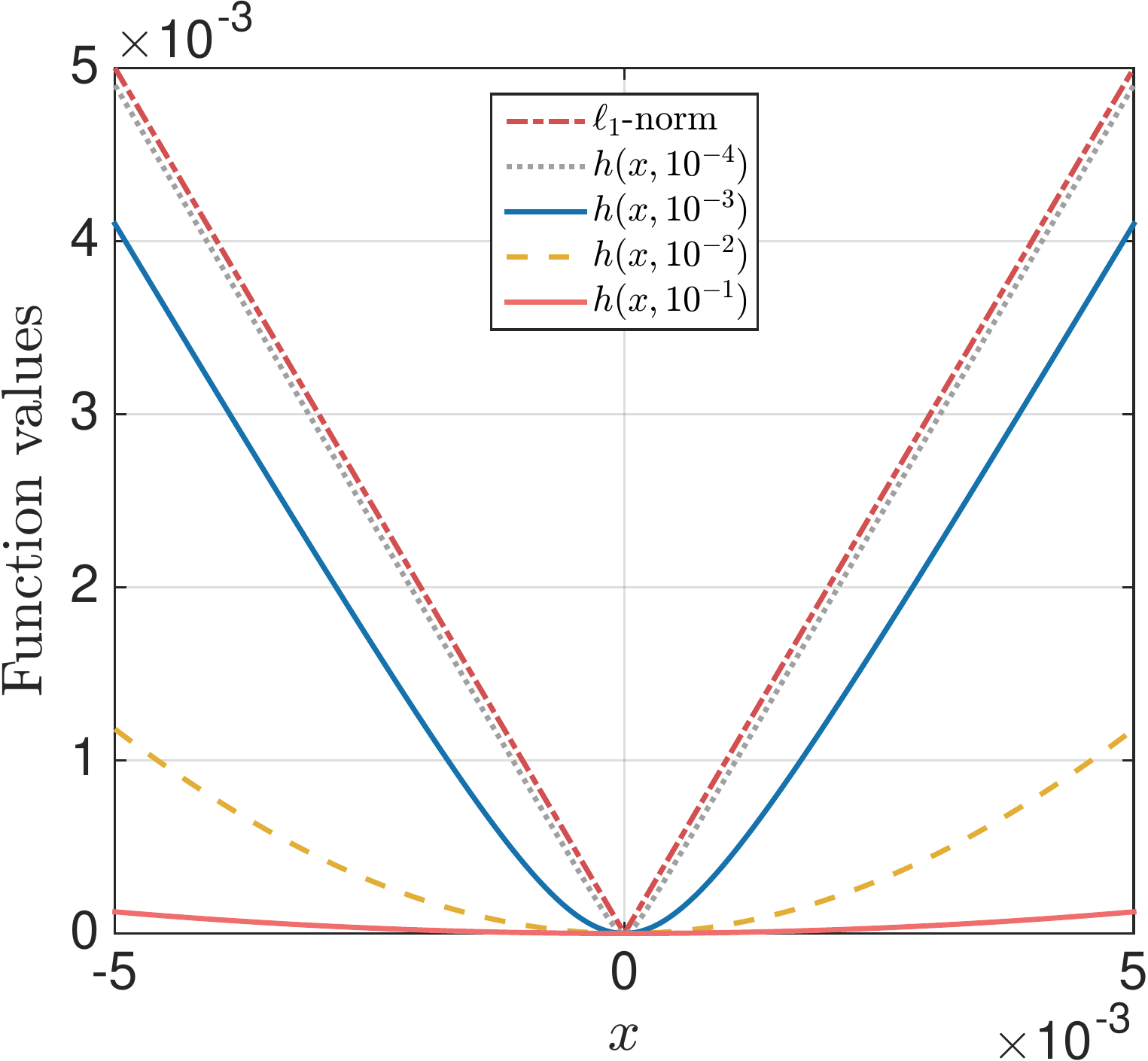}
	\includegraphics[width=0.3\textwidth]{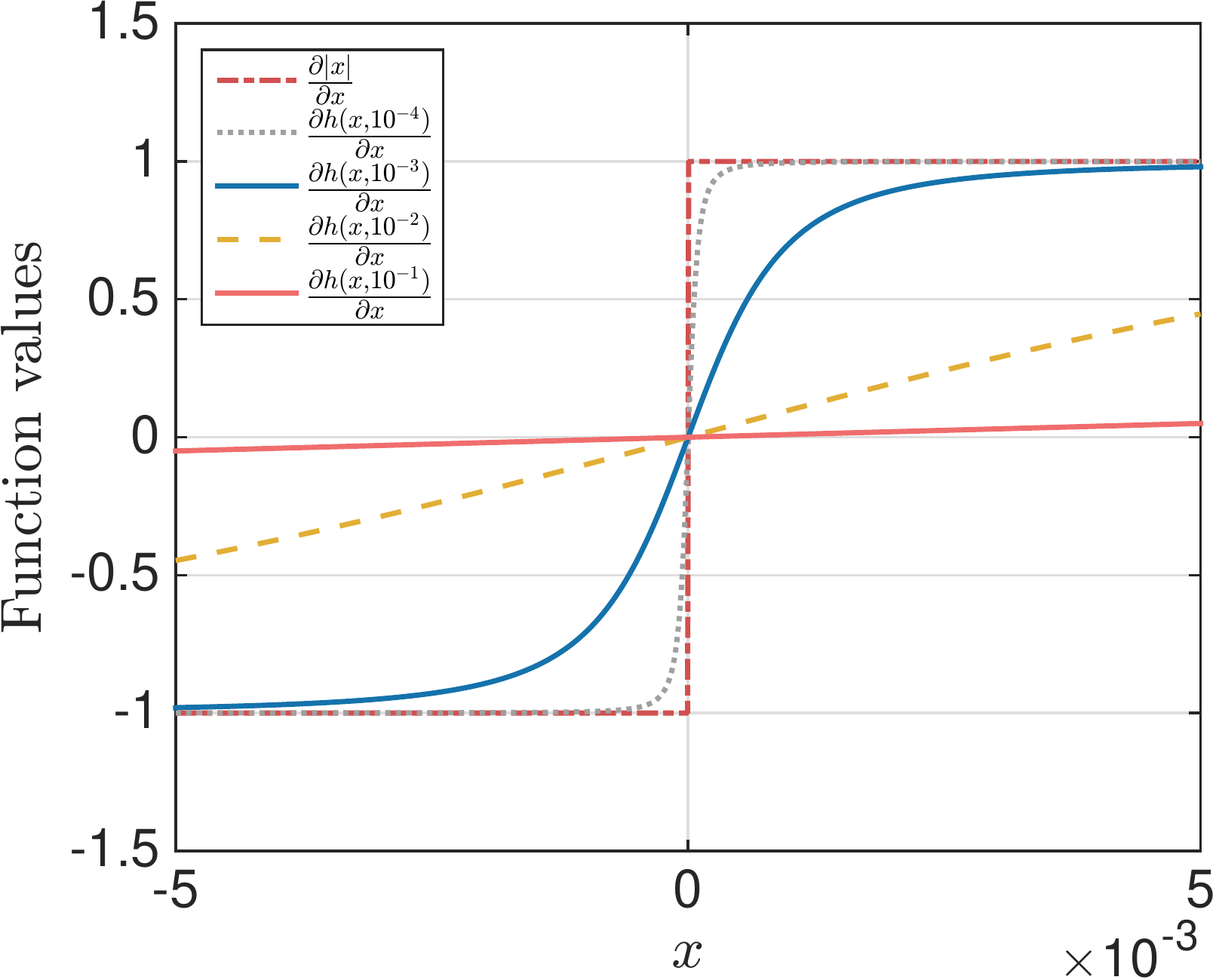}
  \caption{$\ell_1$-norm and its Charbonnier smooth approximations. \textit{Left and middle:} Function values vs. input variable. \textit{Right:} Gradient approximation.} \label{fig:charbonnier}
\end{figure*}

A key assumption in \algo is that $f$ is at least once \emph{differentiable} and has Lipschitz continuous gradients.
Therefore, to connect \algo with our original objective in \eqref{eq:problem}, we will first approximate both the $\ell_1$ and $\ell_\infty$ entrywise matrix norms by smooth functions that have derivatives at least in two degrees. 
For similar approaches in optimization where non-smooth functions are substituted by smooth ones, we refer to the seminal paper of Nesterov \citep{nesterov2007smoothing} and follow-up works \citep{d2008smooth, kelner2014almost}.

\paragraph{Approximating the entrywise $\ell_1$-norm.}
For the approximation of the $\ell_1$-norm, we will use the Charbonnier loss function \citep{charbonnier1994two, barron2017more}, parameterized as follows:
\begin{align}{\label{eq:charbonnier}}
h(x, \tau) = \tau \cdot \left(\sqrt{\left(\frac{x}{\tau}\right)^2 +1} - 1 \right).
\end{align}
To illustrate how a good approximation is \eqref{eq:charbonnier} to the $\ell_1$-norm, see Figure \ref{fig:charbonnier}.

We now discuss about the matrix form of \eqref{eq:charbonnier} and its properties.
With a slight overload of notation, we define the matrix version of \eqref{eq:charbonnier} as follows:
\begin{small}
\begin{align}\label{eq:charbonnier2}
h(X, \tau) &= \sum_{i = 1}^m \sum_{j = 1}^n h(X_{ij}, \tau) \nonumber \\ &:= \tau \cdot \sum_{i = 1}^m \sum_{j = 1}^n \left(\sqrt{\left( \frac{X_{ij}}{\tau}\right)^2 +1} - 1\right).
\end{align}
\end{small}
The distinction between scalar and matrix $h$ will be apparent from the text.
Gradient and Hessian information of $h$ satisfy the following lemma; the proof is deferred to the supp. material:
\begin{lemma}\label{lemma:charbonnier}
For any $X \in \R^{m \times n}$:
\begin{itemize}[leftmargin = 0.4cm]
	\item $\nabla h(X, \tau) = \frac{1}{\tau} X \odot S \in \R^{m \times n}$, where $S \in \R^{m \times n}$ and $S_{ij} := \tfrac{2}{\sqrt{\left(\sfrac{X_{ij}}{\tau}\right)^2 + 1}}$, 
	\item $\nabla^2 h(X, \tau) = \frac{1}{\tau} I \odot Q \in \R^{mn \times mn}$, where $Q \in \R^{mn \times mn}$ and $Q_{ij} := \tfrac{2}{\left(\left(\sfrac{X_{ij}}{\tau}\right)^2 + 1\right)^{3/2}}$.
\end{itemize}
\end{lemma}

The above lead to the following lemma; the proof is provided in the supp. material:
\begin{lemma}{\label{lemma:charbonnier2}}
Function $h$ is a convex continuously differentiable function and it has Lipschitz continuous gradients with constant $\frac{2}{\tau}$. 
Moreover:
\begin{align*}
|X|_1 - mn\tau \leq h(X, \tau) \leq |X|_1.
\end{align*}
\end{lemma}

An alternative to the Charbonnier approximation is the Huber loss function with parameter $\tau$ \citep{huber1964robust}:
\begin{align}
h(x, \tau) = \begin{cases}
      x^2/ 2\tau, & \text{if $|x| \leq \tau$} \\
      |x| - \tau/2, & \text{otherwise}.
    \end{cases} \label{eq:huber}
\end{align} 
Huber loss combines a $\ell_2$-norm measure for small values of $x$ and a $\ell_1$-norm like measure for large $x$. 
Observe in \eqref{eq:huber} that it is only first-order differentiable; thus any computations involving second order derivatives cannot be applied.
On the other hand, the Charbonnier loss function, which is also known as the ``pseudo-Huber loss function", is a smooth approximation of the Huber loss that ensures that derivatives are continuous for all degrees.
W.l.o.g., we focus on the Charbonnier function.

\paragraph{Approximating the entrywise $\ell_\infty$-norm.}
Following similar procedure for the entrywise matrix $\ell_\infty$-norm, we will use the \texttt{logsumexp} function, defined as follows:
\begin{small}
\begin{align}\label{eq:softmax}
\sigma(X, \tau) = \tau \cdot \log \left(\frac{\sum_{i = 1}^m \sum_{j = 1}^n e^{\sfrac{X_{ij}}{\tau}} + e^{\sfrac{-X_{ij}}{\tau}}}{2 m n} \right)
\end{align}
\end{small}
Define matrices $P, N \in \R^{m \times n}$ such that:
$P_{ij} = e^{\sfrac{X_{ij}}{\tau}} + e^{-\sfrac{X_{ij}}{\tau}}$ and 
$N_{ij} = e^{\sfrac{X_{ij}}{\tau}} - e^{-\sfrac{X_{ij}}{\tau}}$.
Then, the following lemma defines the gradient and Hessian information of the \texttt{logsumexp} function; see also the supp. material:
\begin{lemma}\label{lemma:softmax}
For any $X \in \R^{m \times n}$:
\begin{itemize}[leftmargin = 0.4cm]
	\item $\nabla \sigma(X, \tau) = \frac{1}{{\rm Tr}(\mathbb{1} \cdot P)} \cdot N \in \R^{m \times n}$,
	\item \small{$\nabla^2 \sigma(X, \tau) = \frac{\left( \texttt{diag}(\texttt{vec}(P)) - \frac{\texttt{vec}(N) \texttt{vec}(N)^\top}{{\rm Tr}(\mathbb{1} \cdot P)} \right)}{\tau \cdot {\rm Tr}(\mathbb{1} \cdot P)}  \in \R^{mn \times mn}$}
\end{itemize}
where $\texttt{diag}(\cdot): \R^{mn} \rightarrow \R^{mn \times mn}$ turns the vector input to a diagonal matrix output, $\texttt{vec}(\cdot): \R^{m \times n} \rightarrow \R^{mn}$ turns a matrix to a vector by ``stacking" its columns, and $\mathbb{1}$ denotes the all-ones matrix. 
\end{lemma}

Similar to the Charbonnier approximation, we get the following lemma; the proof is in the supp. material:
\begin{lemma}\label{lemma:logsumexp2}
The \texttt{logsumexp} function $\sigma$ is a convex continuously differentiable function and it has Lipschitz continuous gradients with constant $\frac{1}{\tau}$. 
Moreover:
\begin{align*}
|X|_\infty - \tau \log(2 m n) \leq \sigma(X, \tau) \leq |X|_\infty.
\end{align*}
\end{lemma}

\section{An approximate solver for $\ell_p$-norm low rank approximation}{\label{sec:solver}}

The proposed schemes are provided in Algorithms \ref{algo:ell1}-\ref{algo:ellinfty}, and are based on Algorithm \ref{algo:FGD} as a sub-solver.
In order to hope for a good initialization, we consider the smooth versions of \eqref{eq:problem}, as described in Section \ref{sec:func}, with the added twist that we regularize further the objective with a strongly convex component.
\emph{I.e.}, we approximate \eqref{eq:problem} for $p = 1$ with:
\begin{align} \label{eq:approx_problem1}
\min_{U \in \R^{m \times r}, V \in \R^{n \times r}} h(M - UV^\top, \tau) + \frac{\lambda}{2} |UV^\top |_2^2,
\end{align}
and the case $p = \infty$ with
\begin{align} \label{eq:approx_problem2}
\min_{U \in \R^{m \times r}, V \in \R^{n \times r}} \sigma(M - UV^\top, \tau) + \frac{\lambda}{2} |UV^\top|_2^2.
\end{align}
This modification asserts that both \eqref{eq:approx_problem1}-\eqref{eq:approx_problem2} are strongly convex w.r.t. $X$ with parameter $\lambda$; see also the proof of Lemma \ref{lemma:charbonnier2}.
Observe that the smaller the $\lambda$ parameter is, the less the ``drift" from the original problem. 
We remind that the optimal factors of \eqref{eq:problem} are $U^\star$ and $V^\star$, and their product is denoted as $X^\star = U^\star V^{\star \top}$.

\begin{algorithm}[!h]
	\caption{$\ell_1$-norm low rank approximation solver} \label{algo:ell1}
	\begin{algorithmic}[1]		
		\STATE \textbf{Parameters:} $r$, {\rm OPT}, values of $|X^\star|_2^2$ and $\sigma_r(\widehat{X}^\star)$, $\varepsilon > 0$. 
		\STATE Set $\tau = \frac{\varepsilon \cdot {\rm OPT}}{3 m n}$.
		\STATE Set function $T = O\left(\frac{\sigma_r(\widehat{X}_r^\star)}{\varepsilon {\rm OPT}} \right)$.
		\STATE Set $\lambda = \frac{2 \varepsilon \cdot {\rm OPT}}{3 |X^\star|_2^2}$
		\STATE Compute $\widehat{L} = (\frac{1}{\tau} + \lambda)$.
		\STATE Set $f(UV^\top) := h(M - UV^\top, \tau) + \frac{\lambda}{2} |UV^\top|_2^2$.
		\STATE Run Algorithm \ref{algo:FGD} $\left(U_T,~ V_T\right) = \algo(r, T, \tfrac{1}{4}, 1, \widehat{L})$.
	\end{algorithmic}
\end{algorithm}

Let us first focus on the case of $\ell_1$-norm and Algorithm \ref{algo:ell1}.
The following theorem states that, under proper configuration of algorithm's hyperparameters, one can achieve $(1 + \varepsilon)$-OPT approximation guarantee. 
\begin{theorem}{\label{thm:main1}}
Let $\widehat{X} = U_T V_T^\top \in \R^{m \times n}$ be the solution of Algorithm \ref{algo:ell1}. 
Let the optimal function value of \eqref{eq:problem} for $p = 1$ be denoted as ${{\rm OPT}} := \min_{U, V} |M - UV^\top|_1$ and assumed known, or at least be approximable.
Also, assume we know $\sigma_r(\widehat{X}^\star)$ and $|X^\star|_2^2$.
For user defined parameter $\varepsilon > 0$
and setting the Charbonnier parameter $\tau = \frac{\varepsilon \cdot {\rm OPT}}{3 m n}$, and the strong convexity parameter as $\lambda = \frac{2 \varepsilon \cdot {\rm OPT}}{3 |X^\star|_2^2}$, the pair $(U_T, V_T)$ of Algorithm \ref{algo:ell1} satisfies:
\begin{align*}
|M - U_T V_T^\top |_1 \leq (1 + \varepsilon) \cdot {\rm OPT},
\end{align*}
after $T = O\left(\sigma_r(\widehat{X}_r^\star)\left( \tfrac{mn}{\left( \varepsilon {\rm OPT} \right)^2} + \tfrac{1}{\| X^\star \|_2^2}\right)  \right)$ iterations.
\end{theorem}

The proof is provided in the appendix.
In the case where OPT is only approximable, straightforward modifications lead to similar performance (where higher number of iterations required).

\emph{Analytical complexity:} Let us denote the time to compute $\nabla f(\cdot)$ as $t_{{\rm grad}}$. 
The initialization complexity of Algorithm \ref{algo:FGD}, as well as its per iteration complexity, is $O(t_{\rm grad} + m n r)$, where the last term is due to either low-rank SVD calculation or matrix-matrix multiplication.
Running Algorithm \ref{algo:FGD} for $T = O\left(\frac{\sigma_r(\widehat{X}_r^\star)}{\varepsilon {\rm OPT}} \right)$ iterations leads to an overall $O\left(\frac{\sigma_r(\widehat{X}_r^\star)}{\varepsilon {\rm OPT}} \cdot \left(t_{\rm grad} + m n r \right)\right)$ time complexity.

Similarly for the case of $p = \infty$, we use the \texttt{logsumexp} function in Algorithm \ref{algo:ellinfty} to smooth the objective, and we obtain the following guarantees:

\begin{algorithm}[!h]
	\caption{$\ell_\infty$-norm low rank approximation solver} \label{algo:ellinfty}
	\begin{algorithmic}[1]		
		\STATE \textbf{Parameters:} $r$, {\rm OPT}, values of $|X^\star|_2^2$ and $\sigma_r(\widehat{X}^\star)$, $\varepsilon > 0$. 
		\STATE Set $\tau = \frac{\varepsilon \cdot {\rm OPT}}{3 \log (2 m n)}$.
		\STATE Set function $T = O\left(\frac{\sigma_r(\widehat{X}_r^\star)}{\varepsilon {\rm OPT}} \right)$.
		\STATE Set $\lambda = \frac{2 \varepsilon \cdot {\rm OPT}}{3 |X^\star|_2^2}$
		\STATE Compute $\widehat{L} = (\frac{1}{\tau} + \lambda)$.
		\STATE Set $f(UV^\top) := \sigma(M - UV^\top, \tau) + \frac{\lambda}{2} |UV^\top|_2^2$.
		\STATE Run Algorithm \ref{algo:FGD} $(U_T,~ V_T) = \algo(r, T, \tfrac{1}{4}, 1, \widehat{L})$.
	\end{algorithmic}
\end{algorithm}

\begin{corollary}{\label{thm:main2}}
Let $\widehat{X} = U_T V_T^\top \in \R^{m \times n}$ be the solution of Algorithm \ref{algo:ell1}. 
Let the optimal function value of \eqref{eq:problem} for $p = \infty$ be denoted as ${{\rm OPT}} := \min_{U, V} |M - UV^\top|_\infty$, and assumed known, or be at least approximable.
Also, assume we know $\sigma_r(\widehat{X}^\star)$ and $|X^\star|_2^2$.
For user defined approximation parameter $\varepsilon > 0$
and setting the \texttt{logsumexp} parameter $\tau = \frac{\varepsilon \cdot {\rm OPT}}{3 \log (2mn)}$, and the strong convexity parameter as $\lambda = \frac{2 \varepsilon \cdot {\rm OPT}}{3 |X^\star|_2^2}$, the pair $(U_T, V_T)$ of Algorithm \ref{algo:ell1} satisfies:
\begin{align*}
|M - U_T V_T^\top |_\infty \leq (1 + \varepsilon) \cdot {\rm OPT},
\end{align*}
after $T = O\left(\sigma_r(\widehat{X}_r^\star)\left( \tfrac{\log (mn)}{\left( \varepsilon {\rm OPT} \right)^2} + \tfrac{1}{\| X^\star \|_2^2}\right)  \right)$ iterations.
\end{corollary}
Similar analytical complexity can be derived for Algorithm \ref{algo:ellinfty} and is omitted due to lack of space.

Results of similar flavor (and under similar assumptions) can be found in \citep{kelner2014almost} for the problem of maximum flow.
There, the authors consider non-Euclidean gradient descent algorithms for the minimization of $\ell_\infty$-norm over vectors, where the gradient step takes into consideration the geometry of the non-smooth objective with the use of \emph{sharp} operators. 
We applied a similar approach for both $p \in \{1, \infty\}$ in our setting; however, the empirical performance was prohibitive to consider a similar approach here (despite the fact that one can still achieve $(1+ \varepsilon)$-optimal approximation guarantees).

Some remarks regarding the above results.
\begin{remark}
{\it Both algorithms require the knowledge of three quantities: {\rm OPT}, $|X^\star |_2^2$ and $\sigma_r(\widehat{X}^\star)$. 
While finding these values could be as difficult as the original problem \eqref{eq:problem}, these values do not need to be known exactly: 
in particular, the algorithms imply that ``for sufficiently small $\tau$ and $\lambda$ parameters, and for a sufficiently large number of iterations $T$, we can find a good approximation".}
\end{remark}

\begin{remark}
{\it While finding the exact value of {\rm OPT} is difficult, there are problem cases where this value could be easily upper bounded. 
E.g., consider the problem of low-rank matrix approximation from quantization, as noted in \citep{gillis2017low}: 
there, we know from structure that $|M - X^\star|_{\infty} = {\rm OPT} \leq 0.5$.}
\end{remark}

\begin{remark}
{\it Finding a good initialization is a key assumption for Theorem \ref{thm:main1} and its corollary. 
Such assumptions are made also in other non-convex matrix factorization results; see \citep{sun2015guaranteed, zhao2015nonconvex, tu2015low, bhojanapalli2016dropping, park2016non, ge2016matrix, park2016provable, park2016finding, li2016recovery, li2016symmetry, wang2017universal, ge2017no}. 
From \citep{park2016finding}, it is known that we can easily compute such an initialization as the best rank-$r$ approximation of $M$ w.r.t. the $\ell_2$-norm, via SVD. 
In particular, such an initialization satisfies $\dist(\Uinit, \Vinit ; \widehat{X}^\star) \leq \tfrac{\sqrt{2} \cdot \sigma_r(\widehat{X}^\star)^{1/2}}{10\sqrt{\kappa}}$, as long as $f$ is strongly convex with condition number $\kappa \leq 1 + \frac{\sigma_r(\widehat{X}^\star)^2}{4608 \cdot |\widehat{X}_r^\star|_2^2}$. 
While this condition is not easily met in theory (i.e., since $\kappa = \frac{\frac{1}{\tau} + \lambda}{\lambda}$, this means that $\tau$ should be large enough compared to $\lambda$), 
our experiments show that such an initialization performs well.}
\end{remark}

\begin{remark}
{\it As a continuation of the above remark, the reason we use the regularizer $\frac{\lambda}{2} |UV^\top|_2^2$ is to turn the smooth approximations into strongly convex functions (and thus borrow results for initialization). 
In practice, the proposed schemes work as well without the addition of the regularizer; and thus, knowing a priori the quantity $|X^\star |_2^2$ is not necessary in practice.}
\end{remark}

\begin{remark}
\textit{The approach we follow somewhat resembles with the approach proposed in \citep{ke2005robust}. 
There, the authors consider \eqref{eq:problem} for $p = 1$ and propose an alternating minimization scheme. 
Despite the similarities, there are differences with our approach: among which, we perform a single gradient descent step on $U$ and $V$ per iteration, for a smoothed version of \eqref{eq:problem}, instead of minimizing a quadratic programming formulation per each column of $U$ and $V$.
On the contrary, \citep{ke2005robust} handles empirically missing values and weighted low-rank matrix factorization cases; we leave this direction for future research.}
\end{remark}

\section{Experiments}

Our experiments include synthesized applications, in order to highlight the empirical performance of the proposed framework. 
We compare the algorithms in Section \ref{sec:solver} $(i)$ with the algorithms for $\ell_p$-low rank approximation in \citep{chierichetti2017algorithms}, and $(ii)$ with the recent heuristic in \citep{gillis2017low} for $\ell_\infty$-low rank approximation.

Similarly to \citep{chierichetti2017algorithms, gillis2017low} and in order to guarantee fair comparison, we follow in practice the ``folklore" advice for getting an initial estimate for the $\ell_p$-norm problem in \eqref{eq:problem} by beginning with the optimum $\ell_2$-norm solution (\emph{i.e.}, with the low-rank SVD solution). 

\subsection{$\ell_1$-norm approximation}

\begin{figure*}[!ht]
\centering
\includegraphics[width=0.24\textwidth]{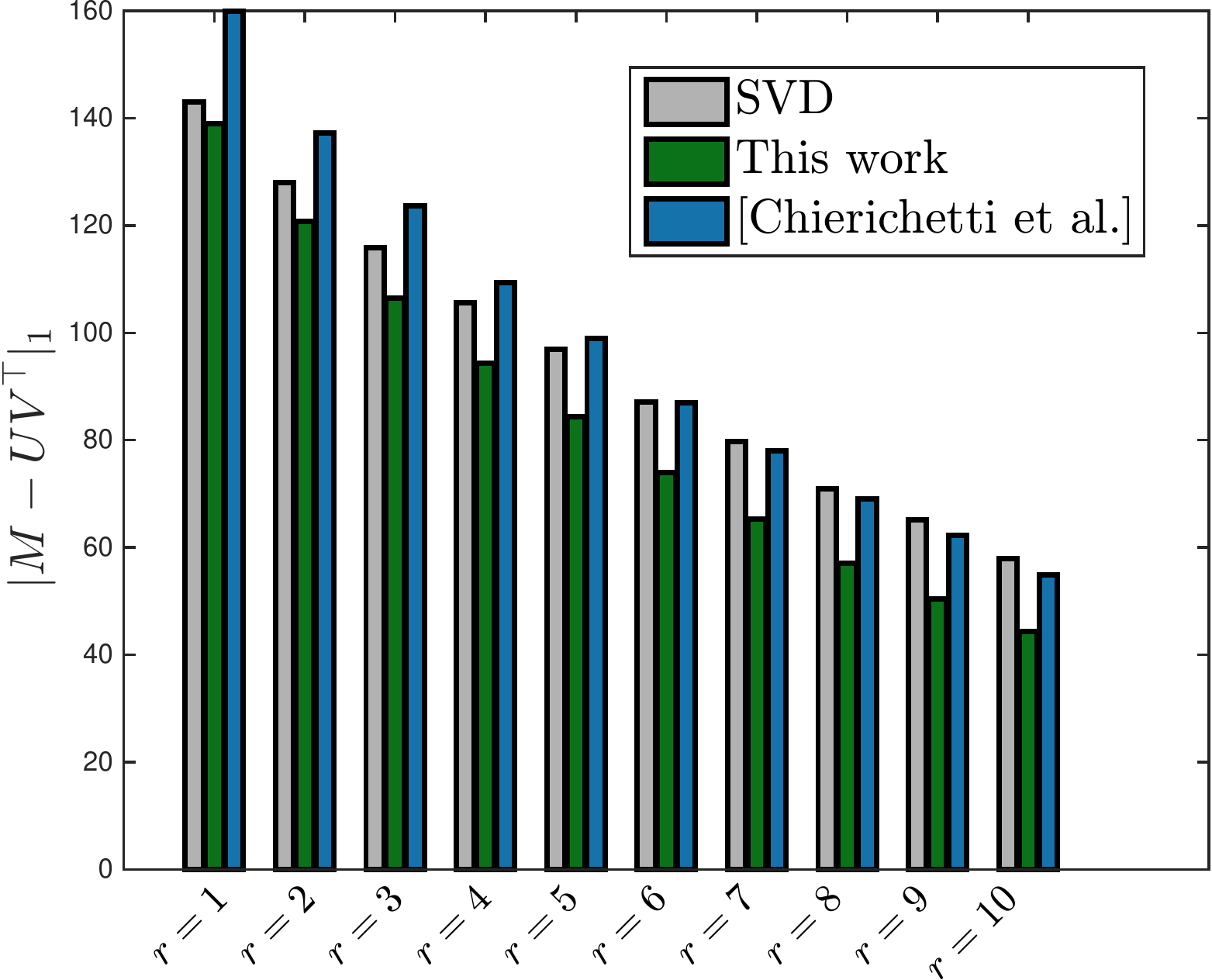} \hspace{-0.2cm}
\includegraphics[width=0.24\textwidth]{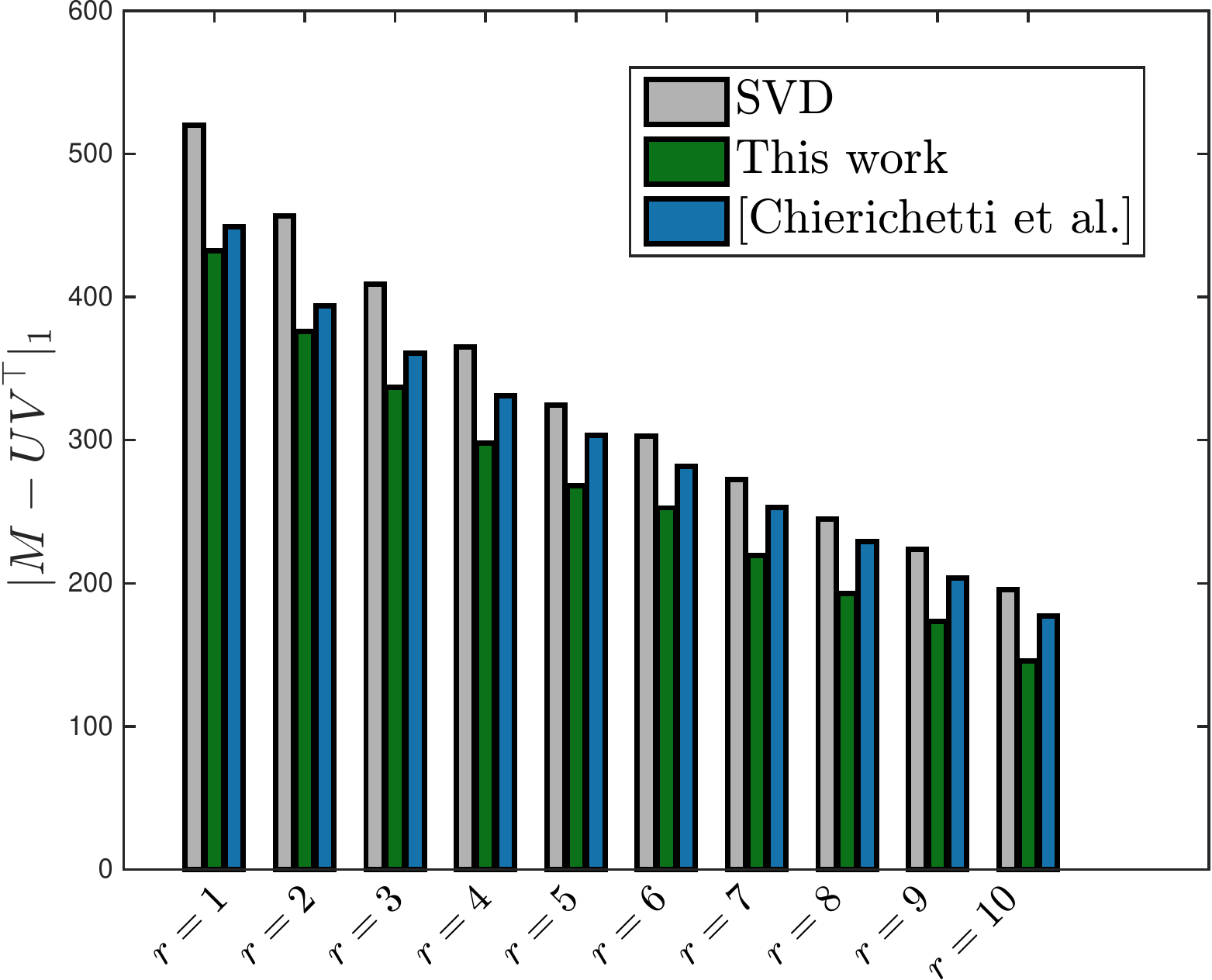} \hspace{-0.2cm}
\includegraphics[width=0.24\textwidth]{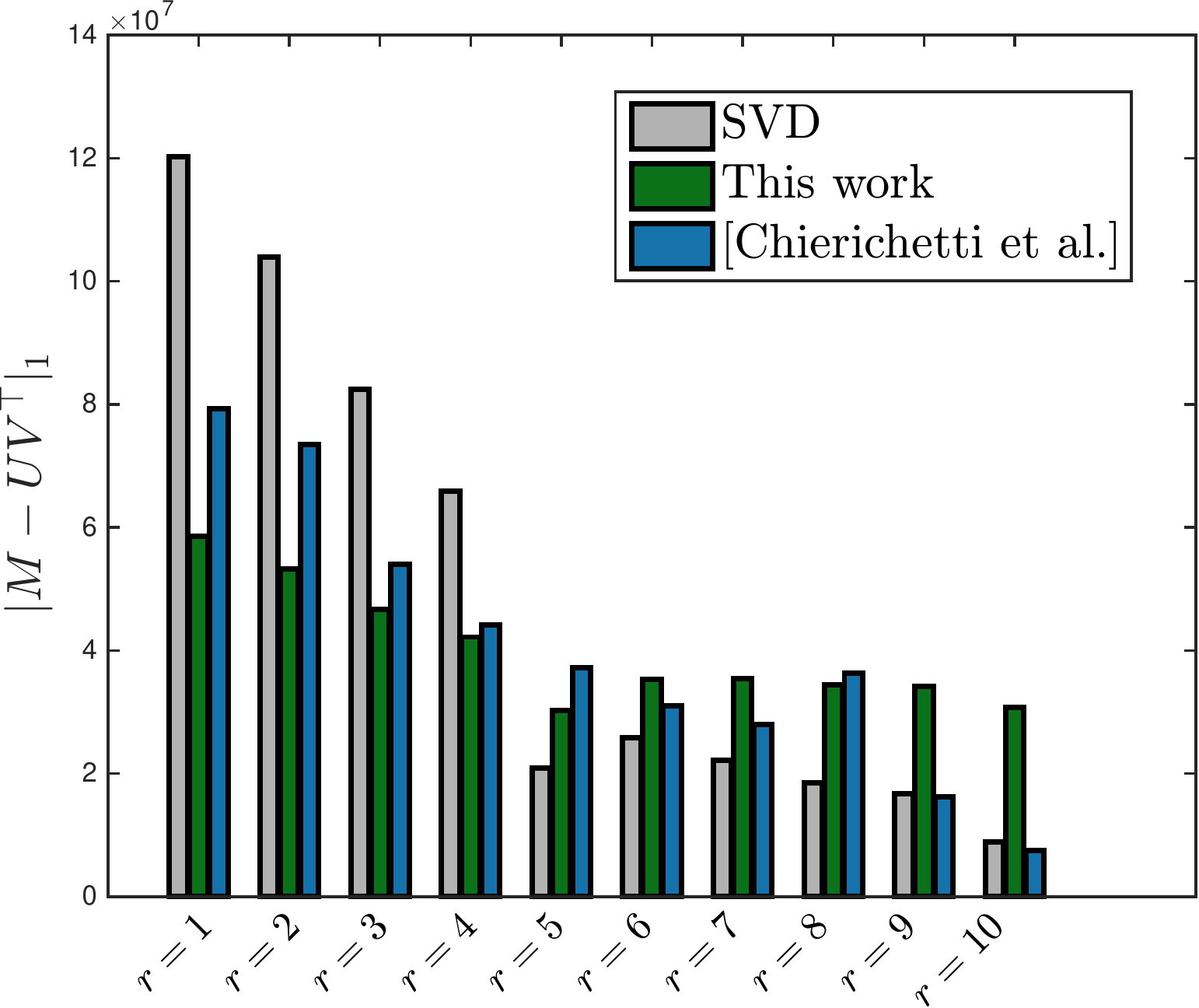} \hspace{-0.2cm}
\includegraphics[width=0.24\textwidth]{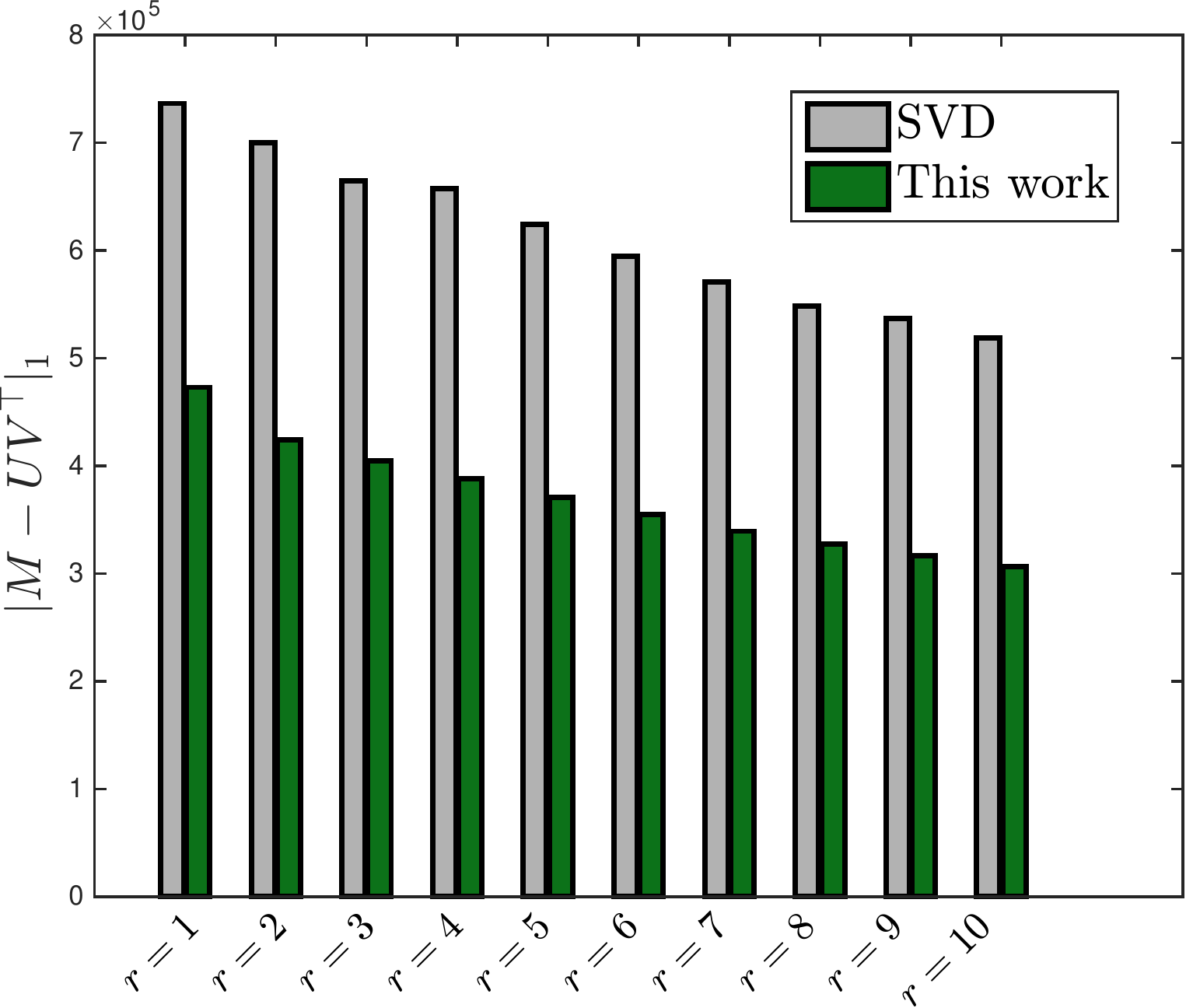} \\
\includegraphics[width=0.24\textwidth]{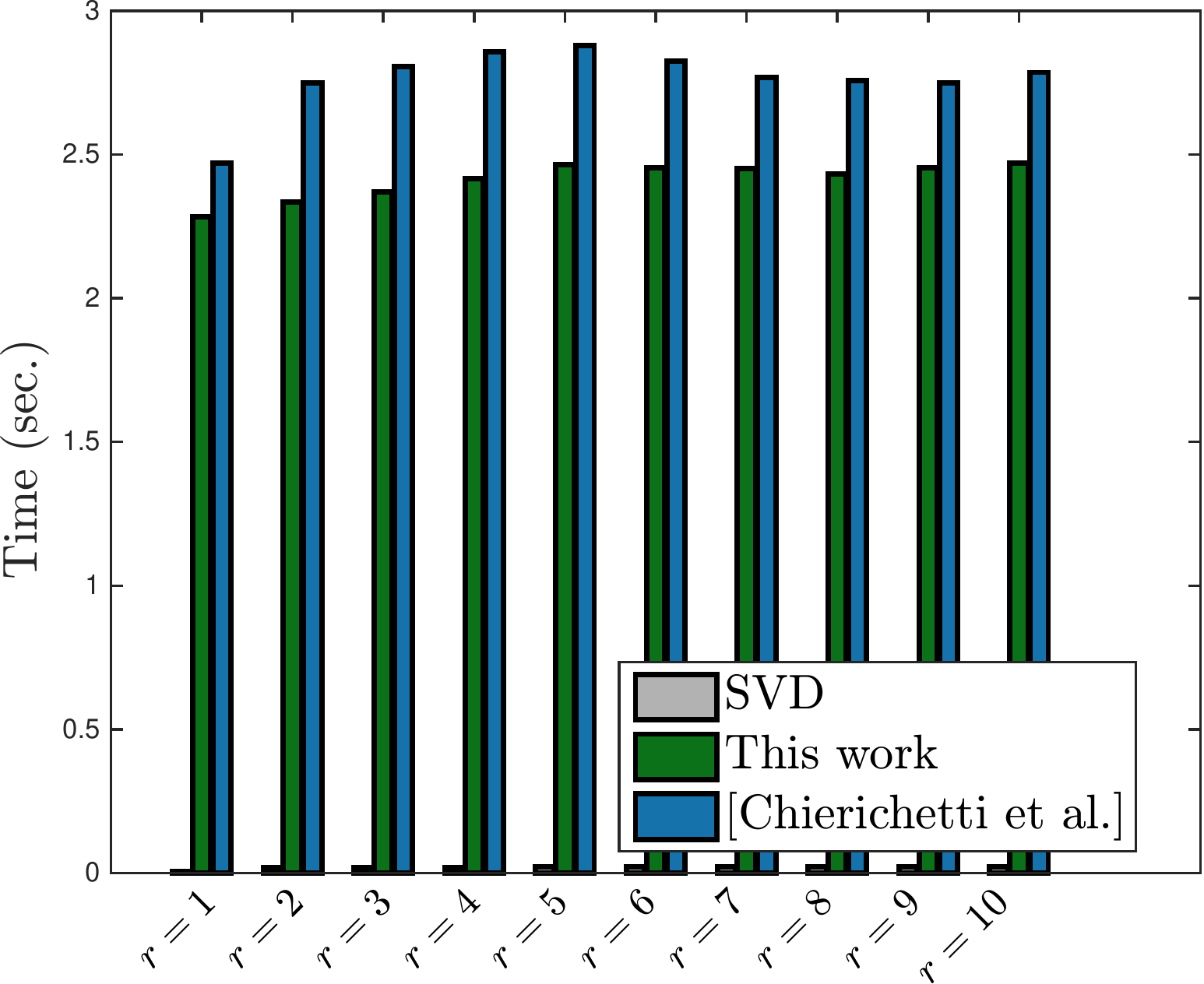} \hspace{-0.2cm}
\includegraphics[width=0.24\textwidth]{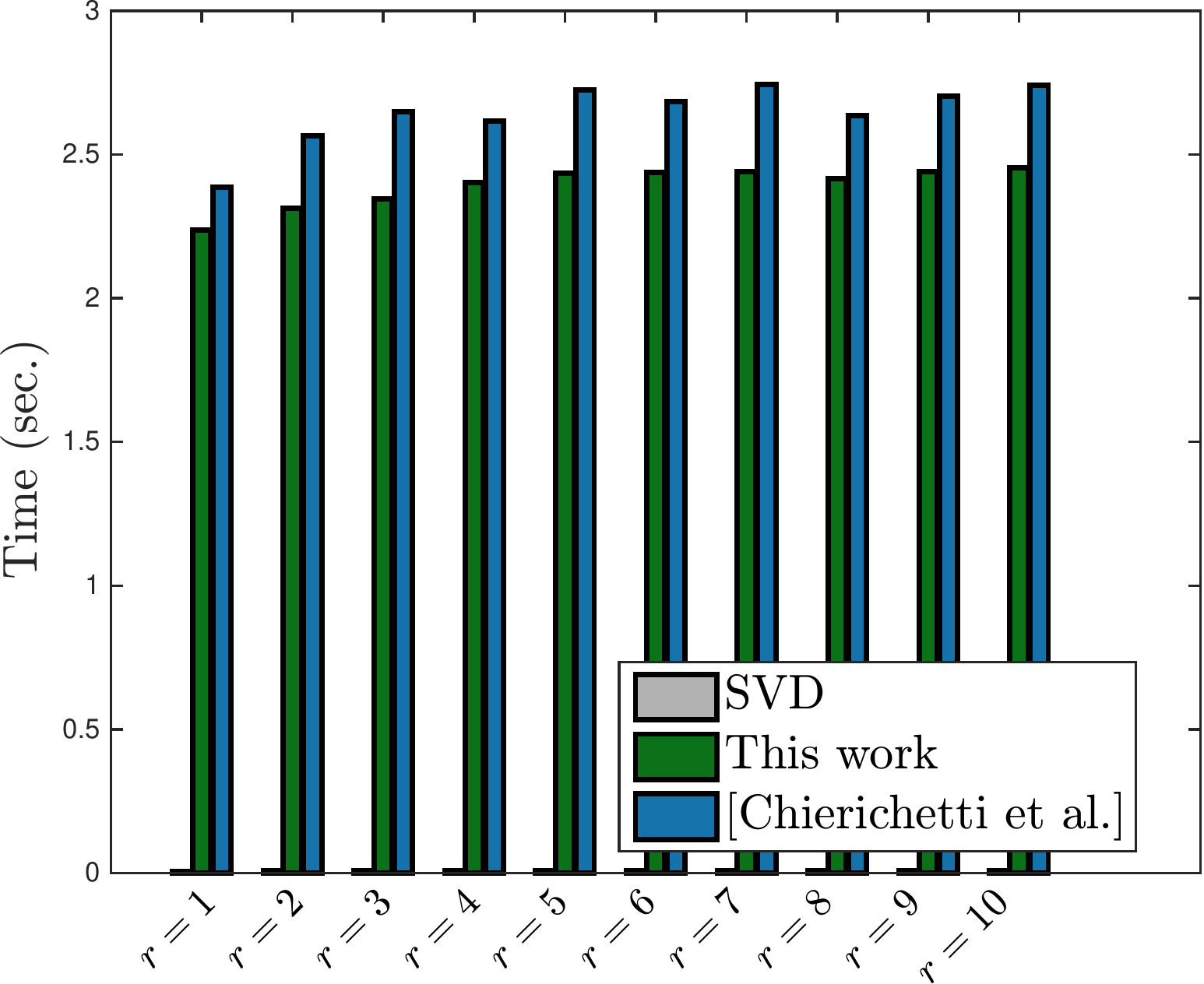} \hspace{-0.2cm}
\includegraphics[width=0.24\textwidth]{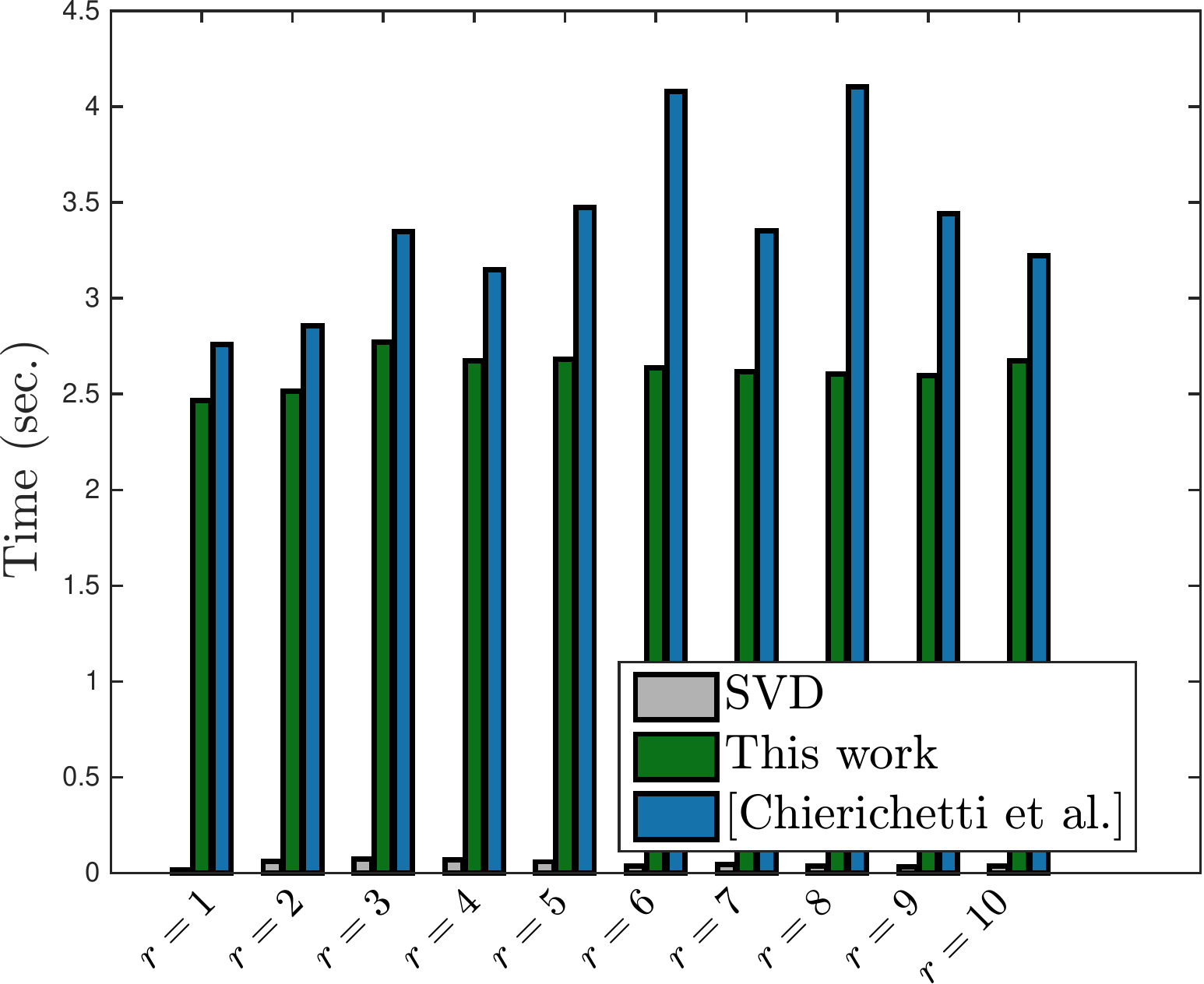} \hspace{-0.2cm}
\includegraphics[width=0.24\textwidth]{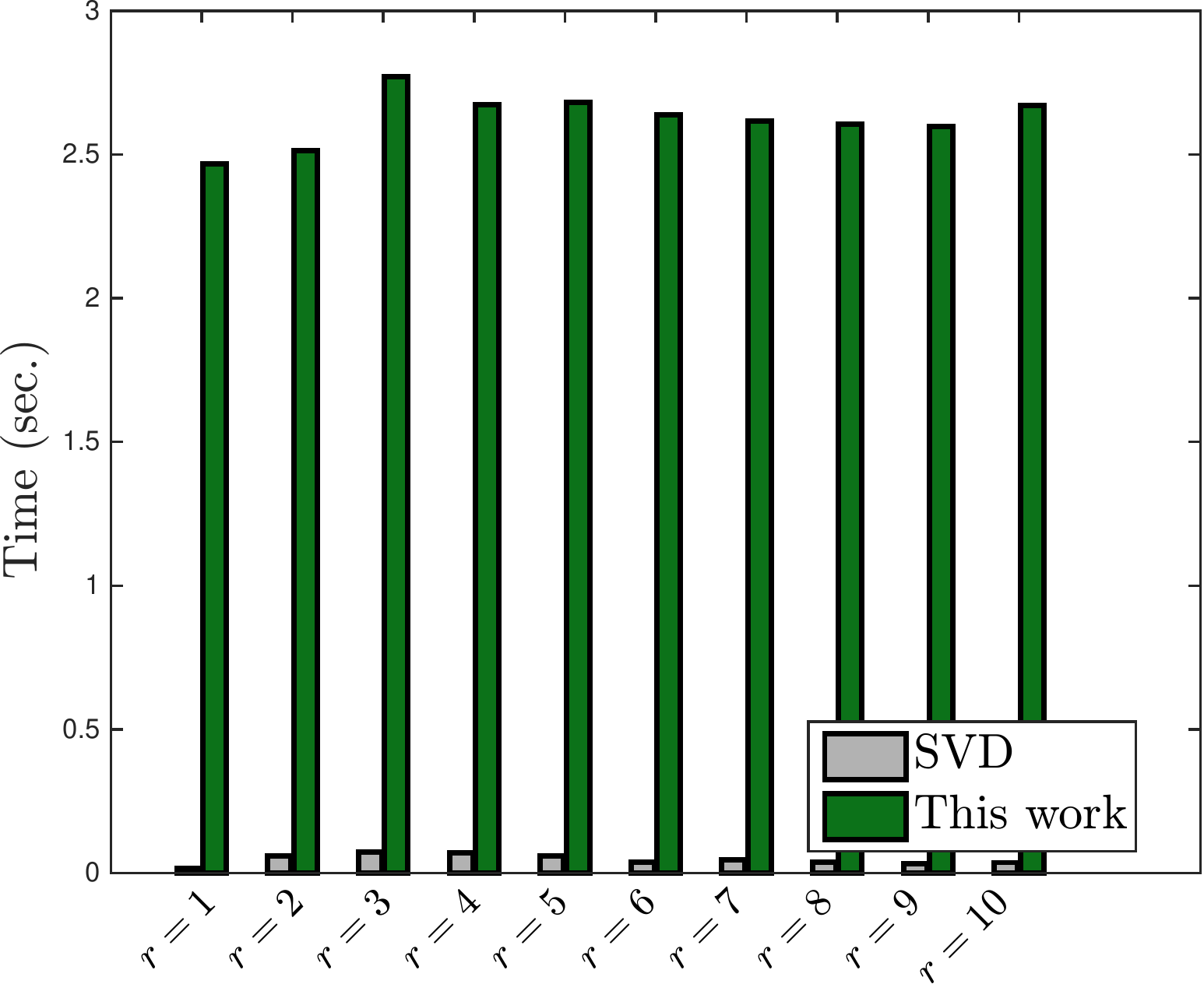}
\caption{Top row: function value performance $|M - UV^\top|_1$; Bottom row: corresponding execution time. In all settings, we set problem \eqref{eq:problem} for $r = \{ 1, \dots, 10\}$. First column: $M \in [0,1]^{20 \times 30}$ where each entry is randomly and independently generated. Second column: $M \in \{-1,1\}^{20 \times 30}$ where each entry is randomly and independently generated. Third column: $M \in \R^{27 \times 27}$ is the FIDAP matrix. Fourth column: $M \in \R^{3430 \times 6906}$ is the word-frequency matrix. In the latter case, the sub-solver for $\ell_p$-projection was not able to complete the task, and thus the algorithm in \citep{chierichetti2017algorithms} is omitted.}
\label{fig:exp1}
\end{figure*}

We perform experiments on both real and synthetic datasets. 
At first, we generate data according to the recent ICML paper \citep{chierichetti2017algorithms}:
We use $20 \times 30$ random matrices $M$, where each entry is a uniformly random value in $[0,1]$. 
Such constructions lead to full rank matrices with high-probability. 
We also construct matrices $M$ of the same size with $\{\pm 1\}$ entries, each selected with $0.5$ probability. 
For real datasets, similar to \citep{chierichetti2017algorithms}, we use
the FIDAP dataset\footnote{\url{http://math.nist.gov/MatrixMarket/data/SPARSKIT/fidap/fidap005.html}} and a word frequency dataset from UC Irvine\footnote{\url{https://archive.ics.uci.edu/ml/datasets/Bag+of+Words}}. 
The FIDAP matrix $M$ is $27 \times 27$ with 279 real asymmetric non-zero entries. 
The word frequency matrix $M$ is $3430 \times 6906$ with $353,160$ non-zero entries.

For the synthesized datasets, we perform $10$ Monte Carlo instantiations and take the median error reported.
For all datasets, we are interested in computing the best rank-$r$ approximation of each $M$ above, w.r.t. the $\ell_1$-norm and for $r \in \{1, \dots, 10\}$.
To compare with \citep{chierichetti2017algorithms}, we use their suggestion and run a simplified version of Algorithm 2 in \citep{chierichetti2017algorithms}, where we repeatedly sample $r$ columns, uniformly at random. 
We then run the $\ell_p$-projection (see Lemma 1 in \citep{chierichetti2017algorithms}) on each sampled set and finally select the solution with the smallest $\ell_p$-error.
For a fair contrast between the algorithms, we first run our algorithm and measure the required time; for approximately the same amount of time, we run \citep{chierichetti2017algorithms}.\footnote{In all our experiments, we make sure the algorithm in \citep{chierichetti2017algorithms} runs at least the same time with our scheme.}
To perform the $\ell_p$-projection, we use \texttt{CVX} package \citep{cvx}.\footnote{We are not aware of another standardized package for $\ell_p$-regression. To accelerate the execution of SeDuMi, we use the lowest precision set up in \texttt{CVX}.}

In our algorithm, we set $\tau = \lambda = 10^{-3}$, and the maximum number of iterations as $T = 4 \cdot 10^4$.
As mentioned above, we use the SVD initialization, and the step size is set according to Algorithm \ref{algo:FGD}.

The results are provided in Figure \ref{fig:exp1}.
Some remarks: $(i)$ for the synthetic cases (two leftmost columns), we observe that our approach attains a better objective function, faster, compared to \citep{chierichetti2017algorithms}. 
Both our work and \citep{chierichetti2017algorithms} is much slower than plain SVD; however, the latter gives a worse solution.
$(ii)$ for the real case (two rightmost columns), our approach is overall better in terms of objective function values; however, this is not universal; there are cases where \citep{chierichetti2017algorithms} (or even SVD) gets to a better result within the same time, especially when $r$ increases. 
For the large matrix case, \citep{chierichetti2017algorithms} with $\texttt{CVX}$ do not scale well; thus omitted.

\begin{table*}[!ht]
\centering
\rowcolors{2}{white}{black!05!white}

\vspace{0.4cm}
\begin{small}
\begin{tabular}{c c c c c}
  \toprule
  & & \multicolumn{3}{c}{\citep{gillis2017low}} \\ 
  & & Time (sec.) & & Error\\
    Rank $r$ & &  \multicolumn{3}{c}{[min, mean, median]} \\  
  \cmidrule{1-1} \cmidrule{3-5} 
 1 & & [6.81e-02, 2.24e-01, 2.28e-01] & & [\textcolor{magenta}{4.91e-01}, \textcolor{magenta}{4.93e-01}, \textcolor{magenta}{4.93e-01}] \\ 
 2 & & [1.55e-02, 2.75e-02, 2.31e-02] & & [5.33e-01, 6.00e-01, 5.96e-01] \\ 
 3 & & [2.42e-02, 5.89e-02, 4.59e-02] & & [5.22e-01, 5.63e-01, 5.44e-01] \\ 
 4 & & [2.69e-02, 4.61e-02, 4.04e-02] & & [5.24e-01, \textcolor{magenta}{5.66e-01}, 5.42e-01] \\ 
 5 & & [4.67e-02, 3.36e-01, 1.48e-01] & & [\textcolor{magenta}{5.04e-01}, 5.36e-01, 5.26e-01] \\ 
 6 & & [6.72e-02, 6.24e-01, 1.34e-01] & & [\textcolor{magenta}{4.98e-01}, 5.20e-01, 5.22e-01] \\ 
 7 & & [5.46e-02, 8.91e-01, 5.47e-01] & & [\textcolor{magenta}{4.90e-01}, \textcolor{magenta}{5.14e-01}, 5.11e-01] \\ 
 8 & & [1.36e-01, 1.66e+00, 5.39e-01] & & [\textcolor{magenta}{4.81e-01}, 5.15e-01, \textcolor{magenta}{5.02e-01}] \\ 
 9 & & [1.90e-01, 2.91e+00, 2.56e+00] & & [\textcolor{magenta}{4.73e-01}, \textcolor{magenta}{4.98e-01}, \textcolor{magenta}{4.89e-01}] \\ 
 10 & & [2.30e-01, 9.60e+00, 4.25e+00] & & [\textcolor{magenta}{4.59e-01}, \textcolor{magenta}{4.97e-01}, 4.79e-01] \\ 
  \bottomrule
\end{tabular}
\end{small}

\vspace{0.4cm}
\begin{small}
\begin{tabular}{c c c c c}
  \toprule
  & & \multicolumn{3}{c}{This work} \\ 
  & & Time (sec.) & & Error\\
  Rank $r$ & & \multicolumn{3}{c}{[min, mean, median]} \\  
  \cmidrule{1-1} \cmidrule{3-5} 
 1 & & [2.57e-02, 4.32e+01, 5.44e+01] & & [4.99e-01, 5.82e-01, 5.01e-01] \\ 
 2 & & [2.60e-02, 4.95e+01, 5.44e+01] & & [\textcolor{magenta}{5.04e-01}, \textcolor{magenta}{5.49e-01}, \textcolor{magenta}{5.07e-01}] \\ 
 3 & & [5.20e+01, 5.43e+01, 5.42e+01] & & [\textcolor{magenta}{5.06e-01}, \textcolor{magenta}{5.10e-01}, \textcolor{magenta}{5.10e-01}] \\ 
 4 & & [1.55e-02, 3.67e+01, 5.15e+01] & & [\textcolor{magenta}{5.05e-01}, 5.90e-01, \textcolor{magenta}{5.10e-01}] \\ 
 5 & & [4.17e-02, 7.92e+01, 8.93e+01] & & [5.07e-01, \textcolor{magenta}{5.33e-01}, \textcolor{magenta}{5.13e-01}] \\ 
 6 & & [7.27e+01, 8.03e+01, 7.76e+01] & & [5.02e-01, \textcolor{magenta}{5.08e-01}, \textcolor{magenta}{5.09e-01}] \\ 
 7 & & [1.62e-02, 5.11e+01, 6.52e+01] & & [5.08e-01, 5.84e-01, \textcolor{magenta}{5.08e-01}] \\ 
 8 & & [5.51e+01, 6.55e+01, 6.73e+01] & & [4.95e-01, \textcolor{magenta}{5.09e-01}, \textcolor{magenta}{5.02e-01}] \\ 
 9 & & [5.36e+01, 5.89e+01, 5.77e+01] & & [4.78e-01, 5.06e-01, 5.06e-01] \\ 
 10 & & [1.69e-02, 3.86e+01, 5.23e+01] & & [4.69e-01, 5.94e-01, \textcolor{magenta}{4.75e-01}] \\ 
   \bottomrule
\end{tabular}
\end{small}
\caption{Attained objective function values and execution time. Table includes minimum, mean and median values for 10 Monte Carlo instances.} \label{table:1}
\end{table*}

\subsection{$\ell_\infty$-norm approximation}
In this experiment, we follow the experimental setting in \citep{gillis2017low}.
We generate matrices $M \in \R^{100 \times 75}$ as follows: 
We generate $\widetilde{M} = U V^\top$ where $U \in \R^{100 \times r}$ and $V \in \R^{75 \times r}$. 
Each $U$ and $V$ is generated i.i.d. from $N(0,1)$.
Given $\widetilde{M}$, we compute the \emph{rounded} version of $\widetilde{M}$ such as $M = \texttt{round}(\widetilde{M})$. 
This procedure guarantees that, given $M$, there is a low-rank matrix $\widetilde{M}$ that satisfies $|M - \widetilde{M}|_{\infty} \leq 0.5$ (since this is an hard problem, this construction gives an idea how far/close we are to a good solution).

We repeat the above procedure for $r = \{1, \dots, 10\}$ and for $10$ Monte Carlo instances.
We report the minimum, mean and median values of the objective function attained and the time required. 
We compare our algorithms with plain SVD and the heuristics in \citep{gillis2017low}.

The results are reported in Table \ref{table:1}.
Our findings show that both our work and the algorithm in \citep{gillis2017low} perform much better (in terms of quality of solution) than plain SVD (the full set of results can be found in the appendix).
Further, the algorithm in \citep{gillis2017low} has time comparable to the implementation of SVD in Matlab, while our proposed algorithm is much slower; accelerating our proposed algorithm is considered future research direction.
However, while our algorithm does not succeed to find solutions with small objective value (see minimum value in table and compare our work with \citep{gillis2017low}), the median value of objective function values over 10 problem instances is lower than that of \citep{gillis2017low}. \emph{I.e.}, the ``typical" achieved objective value is lower than that of \citep{gillis2017low}.\footnote{We ran the algorithm in \citep{gillis2017low} for more time (repeatedly within allowed time) and picked the best minimum result. However, this did not improve the results of \citep{gillis2017low}.}

\section{Conclusion and future work}{\label{sec:future}}
We consider the problem of low-rank matrix approximation, w.r.t. (entrywise) $\ell_p$-norms, and proposed two algorithms that lead to $(1 + \varepsilon)$-OPT approximations.
Our schemes combine ideas from smoothing techniques in convex optimization, as well as recent non-convex gradient descent algorithms.
Key assumption is that problem-related quantities are known or at least are approximable.
Our experiments show that our scheme performs (at least) competitively with state of the art. 

We have provided several possible extensions of this work. 
A particularly interesting open problem is that of weighted low-rank matrix approximation:
\begin{align*}
\min_{U \in \R^{m \times r}, V \in \R^{n \times r}} | W \odot \left( M - UV^\top\right)  |_p,  \quad p \in \{1, \infty\},
\end{align*}
where different assumptions on $W$ lead to different open research questions.

%

\bibliographystyle{plain}
\bibliography{biblio}

\begin{thebibliography}{10}

\bibitem{aanaes2002robust}
H.~Aanas, R.~Fisker, K.~Astrom, and J.~Carstensen.
\newblock Robust factorization.
\newblock {\em IEEE Transactions on Pattern Analysis and Machine Intelligence},
  24(9):1215--1225, 2002.

\bibitem{asteris2016bipartite}
M.~Asteris, A.~Kyrillidis, D.~Papailiopoulos, and A.~Dimakis.
\newblock Bipartite correlation clustering: {M}aximizing agreements.
\newblock In {\em Artificial Intelligence and Statistics}, pages 121--129,
  2016.

\bibitem{barron2017more}
J.~Barron.
\newblock A more general robust loss function.
\newblock {\em arXiv preprint arXiv:1701.03077}, 2017.

\bibitem{bhojanapalli2016dropping}
S.~Bhojanapalli, A.~Kyrillidis, and S.~Sanghavi.
\newblock Dropping convexity for faster semi-definite optimization.
\newblock In {\em 29th Annual Conference on Learning Theory}, pages 530--582,
  2016.

\bibitem{cabral2013unifying}
R.~Cabral, F.~De~la Torre, J.~Costeira, and A.~Bernardino.
\newblock Unifying nuclear norm and bilinear factorization approaches for
  low-rank matrix decomposition.
\newblock In {\em Proceedings of the IEEE International Conference on Computer
  Vision}, 2013.

\bibitem{candes2011robust}
E.~Candes, X.~Li, Y.~Ma, and J.~Wright.
\newblock Robust principal component analysis?
\newblock {\em Journal of the ACM (JACM)}, 58(3):11, 2011.

\bibitem{charbonnier1994two}
P.~Charbonnier, L.~Blanc-Feraud, G.~Aubert, and M.~Barlaud.
\newblock Two deterministic half-quadratic regularization algorithms for
  computed imaging.
\newblock In {\em Image Processing, 1994. Proceedings. ICIP-94., IEEE
  International Conference}, volume~2, pages 168--172. IEEE, 1994.

\bibitem{chiang2016robust}
K.-Y. Chiang, C.-J. Hsieh, and I.~Dhillon.
\newblock Robust principal component analysis with side information.
\newblock In {\em International Conference on Machine Learning}, pages
  2291--2299, 2016.

\bibitem{chierichetti2017algorithms}
F.~Chierichetti, S.~Gollapudi, R.~Kumar, S.~Lattanzi, R.~Panigrahy, and
  D.~Woodruff.
\newblock Algorithms for $\ell_p$ low rank approximation.
\newblock {\em arXiv preprint arXiv:1705.06730}, 2017.

\bibitem{collins2002generalization}
M.~Collins, S.~Dasgupta, and R.~Schapire.
\newblock A generalization of principal components analysis to the exponential
  family.
\newblock In {\em Advances in neural information processing systems}, pages
  617--624, 2002.

\bibitem{csisz1984information}
I.~Csiszar and G.~Tusnady.
\newblock Information geometry and alternating minimization procedures.
\newblock {\em Statistics and decisions}, 1984.

\bibitem{d2008smooth}
A.~d'Aspremont.
\newblock Smooth optimization with approximate gradient.
\newblock {\em SIAM Journal on Optimization}, 19(3):1171--1183, 2008.

\bibitem{eriksson2010efficient}
A.~Eriksson and A.~Van Den~Hengel.
\newblock Efficient computation of robust low-rank matrix approximations in the
  presence of missing data using the $\ell_1$-norm.
\newblock In {\em Computer Vision and Pattern Recognition (CVPR), 2010 IEEE
  Conference on}. IEEE, 2010.

\bibitem{cvx}
Michael G. and Stephen B.
\newblock {CVX}: Matlab software for disciplined convex programming, version
  2.1.
\newblock \url{http://cvxr.com/cvx}, March 2014.

\bibitem{ge2017no}
R.~Ge, C.~Jin, and Y.~Zheng.
\newblock No spurious local minima in nonconvex low rank problems: {A} unified
  geometric analysis.
\newblock {\em arXiv preprint arXiv:1704.00708}, 2017.

\bibitem{ge2016matrix}
R.~Ge, J.~Lee, and T.~Ma.
\newblock Matrix completion has no spurious local minimum.
\newblock {\em To appear in NIPS-16, arXiv preprint arXiv:1605.07272}, 2016.

\bibitem{gillis2017low}
N.~Gillis and Y.~Shitov.
\newblock Low-rank matrix approximation in the infinity norm.
\newblock {\em arXiv preprint arXiv:1706.00078}, 2017.

\bibitem{gillis2014fast}
N.~Gillis and S.~Vavasis.
\newblock Fast and robust recursive algorithmsfor separable nonnegative matrix
  factorization.
\newblock {\em IEEE transactions on pattern analysis and machine intelligence},
  36(4):698--714, 2014.

\bibitem{gillis2015complexity}
N.~Gillis and S.~Vavasis.
\newblock On the complexity of robust {PCA} and $ell_1$-norm low-rank matrix
  approximation.
\newblock {\em arXiv preprint arXiv:1509.09236}, 2015.

\bibitem{golub2012matrix}
G.~Golub and C.~Van~Loan.
\newblock {\em Matrix computations}, volume~3.
\newblock JHU Press, 2012.

\bibitem{gordon2003generalized2}
G.~Gordon.
\newblock Generalized$^2$ linear$^2$ models.
\newblock In {\em Advances in neural information processing systems}, pages
  593--600, 2003.

\bibitem{goreinov2001maximal}
S.~Goreinov and E.~Tyrtyshnikov.
\newblock The maximal-volume concept in approximation by low-rank matrices.
\newblock {\em Contemporary Mathematics}, 280:47--52, 2001.

\bibitem{goreinov2011quasioptimality}
S.~Goreinov and E.~Tyrtyshnikov.
\newblock Quasioptimality of skeleton approximation of a matrix in the
  {C}hebyshev norm.
\newblock In {\em Doklady Mathematics}, volume~83, pages 374--375. Springer,
  2011.

\bibitem{gu2016low}
Q.~Gu, Z.~W. Wang, and H.~Liu.
\newblock Low-rank and sparse structure pursuit via alternating minimization.
\newblock In {\em Artificial Intelligence and Statistics}, pages 600--609,
  2016.

\bibitem{huber1964robust}
P.~Huber.
\newblock Robust estimation of a location parameter.
\newblock {\em The Annals of Mathematical Statistics}, 35(1):73--101, 1964.

\bibitem{ke2003robust}
Q.~Ke and T.~Kanade.
\newblock Robust subspace computation using $\ell_1$-norm.
\newblock 2003.

\bibitem{ke2005robust}
Q.~Ke and T.~Kanade.
\newblock Robust $\ell_1$ factorization in the presence of outliers and missing
  data by alternative convex programming.
\newblock In {\em Computer Vision and Pattern Recognition, 2005. CVPR 2005.
  IEEE Computer Society Conference on}, volume~1, pages 739--746. IEEE, 2005.

\bibitem{kelner2014almost}
J.~Kelner, Y.~T. Lee, L.~Orecchia, and A.~Sidford.
\newblock An almost-linear-time algorithm for approximate max flow in
  undirected graphs, and its multicommodity generalizations.
\newblock In {\em Proceedings of the twenty-fifth annual ACM-SIAM symposium on
  Discrete algorithms}, pages 217--226. SIAM, 2014.

\bibitem{kim2015efficient}
E.~Kim, M.~Lee, C.-H. Choi, N.~Kwak, and S.~Oh.
\newblock Efficient $\ell_1$-norm-based low-rank matrix approximations for
  large-scale problems using alternating rectified gradient method.
\newblock {\em IEEE transactions on neural networks and learning systems},
  26(2):237--251, 2015.

\bibitem{kwak2008principal}
N.~Kwak.
\newblock Principal component analysis based on $\ell_1$-norm maximization.
\newblock {\em IEEE transactions on pattern analysis and machine intelligence},
  30(9):1672--1680, 2008.

\bibitem{kyrillidis2012matrix}
A.~Kyrillidis and V.~Cevher.
\newblock Matrix {ALPS}: {A}ccelerated low rank and sparse matrix
  reconstruction.
\newblock In {\em Statistical Signal Processing Workshop (SSP), 2012 IEEE},
  pages 185--188. IEEE, 2012.

\bibitem{kyrillidis2014matrix}
A.~Kyrillidis and V.~Cevher.
\newblock Matrix recipes for hard thresholding methods.
\newblock {\em Journal of mathematical imaging and vision}, 48(2):235--265,
  2014.

\bibitem{kyrillidis2017provable}
A.~Kyrillidis, A.~Kalev, D.~Park, S.~Bhojanapalli, C.~Caramanis, and
  S.~Sanghavi.
\newblock Provable quantum state tomography via non-convex methods.
\newblock {\em arXiv preprint arXiv:1711.02524}, 2017.

\bibitem{li2016symmetry}
X.~Li, Z.~Wang, J.~Lu, R.~Arora, J.~Haupt, H.~Liu, and T.~Zhao.
\newblock Symmetry, saddle points, and global geometry of nonconvex matrix
  factorization.
\newblock {\em arXiv preprint arXiv:1612.09296}, 2016.

\bibitem{li2016recovery}
Y.~Li, Y.~Liang, and A.~Risteski.
\newblock Recovery guarantee of non-negative matrix factorization via
  alternating updates.
\newblock In {\em Advances in Neural Information Processing Systems}, pages
  4987--4995, 2016.

\bibitem{markopoulos2013some}
P.~Markopoulos, G.~Karystinos, and D.~Pados.
\newblock Some options for $\ell_1$-subspace signal processing.
\newblock In {\em Wireless Communication Systems (ISWCS 2013), Proceedings of
  the Tenth International Symposium on}, pages 1--5. VDE, 2013.

\bibitem{markopoulos2014optimal}
P.~Markopoulos, G.~Karystinos, and D.~Pados.
\newblock Optimal algorithms for $\ell_1$-subspace signal processing.
\newblock {\em IEEE Transactions on Signal Processing}, 62(19):5046--5058,
  2014.

\bibitem{meng2013cyclic}
D.~Meng, Z.~Xu, L.~Zhang, and J.~Zhao.
\newblock A cyclic weighted median method for $\ell_1$ low-rank matrix
  factorization with missing entries.
\newblock In {\em AAAI}, volume~4, page~6, 2013.

\bibitem{nesterov2007smoothing}
Y.~Nesterov.
\newblock Smoothing technique and its applications in semidefinite
  optimization.
\newblock {\em Mathematical Programming}, 110(2):245--259, 2007.

\bibitem{park2016provable}
D.~Park, A.~Kyrillidis, S.~Bhojanapalli, C.~Caramanis, and S.~Sanghavi.
\newblock Provable {B}urer-{M}onteiro factorization for a class of
  norm-constrained matrix problems.
\newblock {\em arXiv preprint arXiv:1606.01316}, 2016.

\bibitem{park2016finding}
D.~Park, A.~Kyrillidis, C.~Caramanis, and S.~Sanghavi.
\newblock Finding low-rank solutions to matrix problems, efficiently and
  provably.
\newblock {\em arXiv preprint arXiv:1606.03168}, 2016.

\bibitem{park2016non}
D.~Park, A.~Kyrillidis, C.~Caramanis, and S.~Sanghavi.
\newblock Non-square matrix sensing without spurious local minima via the
  {B}urer-{M}onteiro approach.
\newblock {\em arXiv preprint arXiv:1609.03240}, 2016.

\bibitem{poljak1993checking}
S.~Poljak and J.~Rohn.
\newblock Checking robust nonsingularity is {NP}-hard.
\newblock {\em Mathematics of Control, Signals, and Systems (MCSS)}, 6(1):1--9,
  1993.

\bibitem{qiu2014recursive}
C.~Qiu, N.~Vaswani, B.~Lois, and L.~Hogben.
\newblock Recursive robust {PCA} or recursive sparse recovery in large but
  structured noise.
\newblock {\em IEEE Transactions on Information Theory}, 60(8):5007--5039,
  2014.

\bibitem{singh2008unified}
A.~Singh and G.~Gordon.
\newblock A unified view of matrix factorization models.
\newblock In {\em Joint European Conference on Machine Learning and Knowledge
  Discovery in Databases}, pages 358--373. Springer, 2008.

\bibitem{song2017low}
Z.~Song, D.~Woodruff, and P.~Zhong.
\newblock Low rank approximation with entrywise $\ell_1$-norm error.
\newblock In {\em Proceedings of the 49th Annual ACM SIGACT Symposium on Theory
  of Computing}, pages 688--701. ACM, 2017.

\bibitem{sun2015guaranteed}
R.~Sun and Z.-Q. Luo.
\newblock Guaranteed matrix completion via nonconvex factorization.
\newblock In {\em {IEEE} 56th Annual Symposium on Foundations of Computer
  Science, {FOCS} 2015}, pages 270--289, 2015.

\bibitem{tippingb1999probabilistic}
M.~Tipping.
\newblock Probabilistic visualisation of high-dimensional binary data.
\newblock In {\em Advances in neural information processing systems}, pages
  592--598, 1999.

\bibitem{tipping1999probabilistic}
M.~Tipping and C.~Bishop.
\newblock Probabilistic principal component analysis.
\newblock {\em Journal of the Royal Statistical Society: Series B (Statistical
  Methodology)}, 61(3):611--622, 1999.

\bibitem{tran2016extended}
Q.~Tran-Dinh and Z.~Zhang.
\newblock Extended {G}auss-{N}ewton and {G}auss-{N}ewton-{ADMM} algorithms for
  low-rank matrix optimization.
\newblock {\em arXiv preprint arXiv:1606.03358}, 2016.

\bibitem{tu2015low}
S.~Tu, R.~Boczar, M.~Simchowitz, M.~Soltanolkotabi, and B.~Recht.
\newblock Low-rank solutions of linear matrix equations via {P}rocrustes flow.
\newblock {\em arXiv preprint arXiv:1507.03566}, 2015.

\bibitem{turk1991eigenfaces}
M.~Turk and A.~Pentland.
\newblock Eigenfaces for recognition.
\newblock {\em Journal of cognitive neuroscience}, 3(1):71--86, 1991.

\bibitem{veldt2017correlation}
N.~Veldt, A.~Wirth, and D.~Gleich.
\newblock Correlation clustering with low-rank matrices.
\newblock In {\em Proceedings of the 26th International Conference on World
  Wide Web}, pages 1025--1034. International World Wide Web Conferences
  Steering Committee, 2017.

\bibitem{wang2017universal}
L.~Wang, X.~Zhang, and Q.~Gu.
\newblock A universal variance reduction-based catalyst for nonconvex low-rank
  matrix recovery.
\newblock {\em arXiv preprint arXiv:1701.02301}, 2017.

\bibitem{wiberg1976computation}
T.~Wiberg.
\newblock Computation of principal components when data are missing.
\newblock In {\em Proc. of Second Symp. Computational Statistics}, pages
  229--236, 1976.

\bibitem{xu2010robust}
H.~Xu, C.~Caramanis, and S.~Sanghavi.
\newblock Robust {PCA} via outlier pursuit.
\newblock In {\em Advances in Neural Information Processing Systems}, pages
  2496--2504, 2010.

\bibitem{yi2016fast}
X.~Yi, D.~Park, Y.~Chen, and C.~Caramanis.
\newblock Fast algorithms for robust {PCA} via gradient descent.
\newblock In {\em Advances in neural information processing systems}, pages
  4152--4160, 2016.

\bibitem{zhao2015nonconvex}
T.~Zhao, Z.~Wang, and H.~Liu.
\newblock A nonconvex optimization framework for low rank matrix estimation.
\newblock In {\em Advances in Neural Information Processing Systems}, pages
  559--567, 2015.

\bibitem{zhou2011godec}
T.~Zhou and D.~Tao.
\newblock Go{D}ec: {R}andomized low-rank \& sparse matrix decomposition in
  noisy case.
\newblock In {\em International conference on machine learning}. Omnipress,
  2011.

\end{thebibliography}

\clearpage
\onecolumn
\section{Proofs of lemmata}

\subsection{Proof of Lemma \ref{lemma:charbonnier}}

Due to the decomposability of \eqref{eq:charbonnier2}, we observe $\forall X$:
\begin{align*}
\frac{\partial h(X, \tau)}{\partial X_{ij}} = \frac{2X_{ij}}{\tau} \cdot \left( \left(\frac{X_{ij}}{\tau} \right)^2 + 1\right)^{-1/2} = \frac{X_{ij}}{\tau} \cdot \frac{2}{\sqrt{\left(\frac{X_{ij}}{\tau} \right)^2 + 1}} 
\end{align*} Thus, in compact form, $\nabla h(X, \tau) = \frac{1}{\tau} X \odot S$, where $S$ is defined in the lemma.

Regarding the Hessian information, first observe that $\frac{\partial^2 h(X, \tau)}{\partial X_{ij} \partial X_{lq}} = \frac{\partial \left(\frac{X_{ij}}{\tau} \cdot \frac{2}{\sqrt{\left(\frac{X_{ij}}{\tau} \right)^2 + 1}}\right)}{\partial X_{lq}} = 0$, for indices $(i, j) \neq (l, q)$.
This means that the off-diagonals of $\nabla^2 h(X, \tau)$ are zero.
For the case where $(i, j) = (l, q)$, we have:
\begin{align*}
\frac{\partial^2 h(X, \tau)}{\partial X_{ij}^2} &= \frac{\partial \left(\frac{X_{ij}}{\tau} \cdot \frac{2}{\sqrt{\left(\frac{X_{ij}}{\tau} \right)^2 + 1}}\right)}{\partial X_{ij}} = \frac{2}{\tau} \cdot \frac{\sqrt{(\sfrac{X_{ij}}{\tau})^2 + 1} - \frac{X_{ij}^2}{\tau^2} \cdot \left( \left(\sfrac{X_{ij}}{\tau}\right)^2 + 1\right)^{-1/2}}{\left(\frac{X_{ij}}{\tau}\right)^2 + 1} \\
&= \frac{2}{\tau} \cdot \frac{\left(\sfrac{X_{ij}}{\tau}\right)^2 + 1 - \left(\sfrac{X_{ij}}{\tau}\right)^2}{\left( \left(\frac{X_{ij}}{\tau}\right)^2 + 1\right)^{3/2}} = \frac{1}{\tau} \cdot \frac{2}{\left( \left(\frac{X_{ij}}{\tau}\right)^2 + 1\right)^{3/2}}
\end{align*}
Then, $\nabla^2 h(X, \tau) = \frac{1}{\tau} I \odot Q$, where $Q$ is defined in the lemma.

\subsection{Proof of Lemma \ref{lemma:charbonnier2}}

The first part of the lemma is easily deduced from Lemma \ref{lemma:charbonnier}. 
Observe that $0 \preceq \nabla^2 h(X, \tau) \preceq \frac{2}{\tau} I, ~\forall X$; that is $h$ function is convex with Lipschitz constant $\frac{2}{\tau}$. 
Moreover, by combining $h$ with any strongly convex function $\psi(\cdot)$, say $\psi(X) := \tfrac{\lambda}{2} |X|_2^2$, we easily observe that the composite form $h(X, \tau) + \psi(X)$ satisfies $\lambda I \preceq \nabla^2 h(X, \tau) + \nabla^2 \psi(X) \preceq \left(\frac{2}{\tau} + \lambda \right) I$; \emph{i.e.}, the composite form is also strongly convex.

The last part of the lemma is true because 
\begin{align*}
|X|_1 \geq h(X, \tau) = \sum_{i = 1}^m \sum_{j = 1}^n h(X_{ij}, \tau) &= \tau \cdot \sum_{i = 1}^m \sum_{j = 1}^n \left(\sqrt{\left( \frac{X_{ij}}{\tau}\right)^2 +1} - 1\right) = \sum_{i = 1}^m \sum_{j = 1}^n \left(\sqrt{X_{ij}^2 + \tau^2} - \tau\right) \\ &\geq \sum_{i = 1}^m \sum_{j = 1}^n |X_{ij}| - mn \tau = |X|_1 - mn\tau.
\end{align*}

\subsection{Proof of Lemma \ref{lemma:softmax}}
The proof is elementary as in Lemma \ref{lemma:charbonnier} and we state it for completeness.
First, observe that \eqref{eq:softmax} can be re-written as follows:
\begin{align*}
\sigma(X, \tau) = \tau \cdot \log \left(\frac{{\rm Tr}(\mathbb{1} \cdot P)}{2 m n} \right)
\end{align*}
Observe that calculating gradients with respect to $X_{ij}$, the denominator $2mn$ plays no role.
Following similar motions, we compute partial derivatives as:
\begin{align*}
\frac{\partial \sigma(X, \tau)}{\partial X_{ij}} = \tau \cdot \frac{1}{{\rm Tr}(\mathbb{1} \cdot P)} \cdot \frac{\partial \left(e^{\sfrac{X_{ij}}{\tau}} + e^{-\sfrac{X_{ij}}{\tau}}\right)}{\partial X_{ij}} = \frac{1}{{\rm Tr}(\mathbb{1} \cdot P)} \cdot \left(e^{\sfrac{X_{ij}}{\tau}} - e^{-\sfrac{X_{ij}}{\tau}}\right)
\end{align*}
Gathering all the partial derivatives in a matrix, we get the reported result.

Computing second-order partial derivatives for $\sigma(X, \tau)$, we distinct the cases of diagonal and off-diagonal elements.
For the former, we have:
\begin{align*}
\frac{\partial^2 \sigma(X, \tau)}{\partial X_{ij}^2} = \frac{1}{\tau} \cdot \frac{{\rm Tr}(\mathbb{1} \cdot P) - N_{ij}^2}{{\rm Tr}(\mathbb{1} \cdot P)^2}
\end{align*} 
and for the latter:
\begin{align*}
\frac{\partial^2 \sigma(X, \tau)}{\partial X_{ij} \partial X_{l, q}} = - \frac{1}{\tau} \cdot \frac{- N_{ij} N_{lq}}{{\rm Tr}(\mathbb{1} \cdot P)^2}
\end{align*} 
Combining the two, we get the required result.

\subsection{Proof of Lemma \ref{lemma:logsumexp2}}

Let us first prove convexity. 
By the definition of the Hessian, we want to prove 
$${\rm Tr}(\mathbb{1} \cdot P) \cdot y^\top \left( \texttt{diag}(\texttt{vec}(P)) - \frac{\texttt{vec}(N) \texttt{vec}(N)^\top}{{\rm Tr}(\mathbb{1} \cdot P)}\right) y \geq 0, \quad \quad \forall y \in \R^{mn}.$$
First, observe that ${\rm Tr}(\mathbb{1} \cdot P) \geq 0$ since each element of $P$ is positive by definition.
Second, for $P_{ij} \geq 0, ~\forall i, j$, it is obvious that $\frac{\texttt{vec}(P) \texttt{vec}(P)^\top}{{\rm Tr}(\mathbb{1} \cdot P)} \preceq \texttt{diag}(\texttt{vec}(P))$. 
Thus, what is left is to prove $y^\top \left(\texttt{vec}(N) \texttt{vec}(N)^\top\right) y \leq y^\top \left(\texttt{vec}(P) \texttt{vec}(P)^\top\right) y$, which is true since:
\begin{align*}
y^\top \left(\texttt{vec}(N) \texttt{vec}(N)^\top\right) y &= \|y^\top \texttt{vec}(N) \|_2^2 = \sum_{i = 1}^{mn} (y_i \cdot \texttt{vec}(N)_i)^2 \leq \sum_{i = 1}^{mn} y_i^2 \cdot  \texttt{vec}(N)_i^2 \\ 
&\leq \sum_{i = 1}^{mn} y_i^2 \cdot \texttt{vec}(P)_i^2 = \|y^\top \texttt{vec}(P) \|_2^2 = y^\top \left(\texttt{vec}(P) \texttt{vec}(P)^\top\right) y,
\end{align*}
since $P_{ij} \geq N_{ij}$.
Upper bounding the Hessian, 
\begin{align*}
y^\top \nabla^2 \sigma(X, \tau) y &= y^\top \left(\frac{1}{\tau} \cdot \frac{1}{{\rm Tr}(\mathbb{1} \cdot P)} \cdot \left( \texttt{diag}(\texttt{vec}(P)) - \frac{\texttt{vec}(N) \texttt{vec}(N)^\top}{{\rm Tr}(\mathbb{1} \cdot P)} \right)\right) y \\
 											&\leq y^\top \left(\frac{1}{\tau} \cdot \frac{1}{{\rm Tr}(\mathbb{1} \cdot P)} \cdot \left( \texttt{diag}(\texttt{vec}(P))\right)\right) y \\ 
 											&= \frac{\sum_{i = 1}^{mn} y_i^2 \cdot \texttt{vec}(P)_i}{\tau \cdot {\rm Tr}(\mathbb{1} \cdot P)} \leq \frac{\sum_{i = 1}^{mn} |y_i|^2 \cdot \left(\sum_{i = 1}^{mn} \texttt{vec}(P)_i\right)}{\tau \cdot {\rm Tr}(\mathbb{1} \cdot P)} = \frac{\| y \|_2^2}{\tau}.
\end{align*}
This means that $\sigma$ function is Lipschitz gradient continuous with constant $\frac{1}{\tau}$.
To prove the set of inequalities of the lemma, we observe:
\begin{align*}
 |X|_\infty \geq \sigma(X, \tau) \geq \tau \cdot \log \left( \frac{e^{\sfrac{|X|_\infty}{\tau}}}{2 mn}\right) = |X|_{\infty} - \tau \log (2 m n).
\end{align*}

\subsection{Proof of Theorem \ref{thm:main1}}

Using Lemma \ref{lemma:charbonnier2}, we bound $|M - U_TV_T^\top|_1$ as follows:
\begin{align*}
|M - U_TV_T^\top|_1 &\leq h(M - U_T V_T^\top, \tau) + m n \tau \\
			        &\leq h(M - U_T V_T^\top, \tau) + \frac{\lambda}{2} |U_T V_T^\top|_2^2 + m n \tau 						   
\end{align*}
Define $f: \R^{m \times n} \rightarrow \R$ such as $f(UV^\top) := h(M - UV^\top, \tau) + \frac{\lambda}{2} |UV^\top|_2^2$.
Observe that $f$ is $\lambda$-strongly convex with Lipscihtz continuous gradients with parameter $(\tfrac{2}{\tau} + \lambda)$. 
By Theorem \ref{thm:smooth}, we know that:
$$
f(U_T V_T^\top) - f(\widehat{U}^\star \widehat{V}^{\star \top}) \le \frac{10 \cdot \dist(\Uinit,\Vinit; \widehat{X}^\star_r)^2}{\eta T}.
$$
where $\dist(\Uinit, \Vinit ; \widehat{X}^\star_r) \leq \tfrac{\sqrt{2} \cdot \sigma_r(\widehat{X}^\star_r)^{1/2}}{10\sqrt{\kappa}}$.
Combining this bound with the above, we get:
\begin{align}
|M - U_TV_T^\top|_1 \leq h(M - \widehat{U}^\star \widehat{V}^{\star \top}, \tau) + \frac{\lambda}{2} |\widehat{X}^\star|_2^2 + \frac{10 \cdot \dist(\Uinit,\Vinit;\widehat{X}_r^\star)^2}{\eta T} + mn \tau \label{eq:fix00}
\end{align}
We know from Lemma 4.2 that:
\begin{align*}
h(M - UV^\top, \tau) \leq | M - UV^\top |_1  \quad \Longrightarrow \quad h(M - UV^\top, \tau) + \frac{\lambda}{2} |  UV^\top |_2^2 \leq | M - UV^\top |_1 + \frac{\lambda}{2} | U V^\top |_2^2 
\end{align*}
for every $U,~V$. 
This further implies that:
\begin{align*}
\min_{U, V} \left( h(M - UV^\top, \tau) + \frac{\lambda}{2} |  UV^\top |_2^2\right) &\leq \min_{U, V} \left(| M - UV^\top |_1 + \frac{\lambda}{2} |  UV^\top |_2^2\right) \quad  \Rightarrow \\
h(M - \widehat{U}^\star \widehat{V}^{\star \top}, \tau) + \frac{\lambda}{2} |\widehat{U}^\star \widehat{V}^{\star \top} |_2^2 &\stackrel{(i)}{\leq} \min_{U, V} \left(| M - UV^\top |_1 + \frac{\lambda}{2} |  UV^\top |_2^2\right) \\ & \stackrel{(ii)}{\leq} |M - U^\star V^{\star \top}|_1 + \frac{\lambda}{2} |U^\star V^{\star \top} |_2^2 \quad \\
&\stackrel{(iii)}{=} {\rm OPT} + \frac{\lambda}{2} |U^\star V^{\star \top} |_2^2
\end{align*}
where $(i)$ is due to the optimality of $\widehat{U}^\star, \widehat{V}^\star$ as the minimizer of $f(UV^\top) := h(M - UV^\top, \tau) + \frac{\lambda}{2} |  UV^\top |_2^2$, 
$(ii)$ is due to $U^\star, V^\star$ not being necessarily the minimizers of $\min_{U, V} \left(| M - UV^\top |_1 + \frac{\lambda}{2} |  UV^\top |_2^2\right) $, and 
$(iii)$ ${\rm OPT} := \min_{U, V} | M - UV^\top |_1 = |M - U^\star V^{\star \top}|_1 $.
Thus, \eqref{eq:fix00} becomes:
\begin{align*}
|M - U_TV_T^\top|_1 
&\leq{\rm OPT} + \frac{\lambda}{2} |X^\star|_2^2 + \frac{10 \cdot \dist(\Uinit,\Vinit;\Xoptr)^2}{\eta T} + mn \tau
\end{align*}
For $\varepsilon > 0$, setting $\tau = \frac{\varepsilon \cdot {\rm OPT}}{3 m n}$ we observe that $mn \tau = \frac{\varepsilon \cdot {\rm OPT}}{3}$. 
Executing Algorithm \ref{algo:FGD} for $T \geq \frac{10 \cdot \sigma_r(\widehat{X}_r^\star)}{50} \cdot \frac{3}{\eta \varepsilon {\rm OPT}}$, we can guarantee that $\frac{10 \cdot \dist(\Uinit,\Vinit;\widehat{X}_r^\star)^2}{\eta T} \leq \frac{10 \sigma_r(\widehat{X}^\star)}{50 \eta \cdot \frac{3 \cdot 10 \cdot \sigma_r(\widehat{X}^\star)}{50 \eta \varepsilon {\rm OPT}}} = \frac{\varepsilon \cdot {\rm OPT}}{3}$.
Finally, setting $\lambda = \frac{2 \varepsilon \cdot {\rm OPT}}{3 |X^\star|_2^2}$, we obtain: $\frac{2 \varepsilon \cdot {\rm OPT}}{6 |X^\star|_2^2} \cdot |X^\star|_2^2 = \frac{\varepsilon \cdot {\rm OPT}}{3}$. 
Substituting the above in the main recursion, we get:
\begin{align*}
|M - U_TV_T^\top|_1 &\leq {\rm OPT} + \frac{\lambda}{2} |X^\star|_2^2 + \frac{10 \cdot \dist(\Uinit,\Vinit;\Xoptr)^2}{\eta T} + mn \tau \\
								&\leq {\rm OPT} + \frac{\varepsilon \cdot {\rm OPT}}{3} + \frac{\varepsilon \cdot {\rm OPT}}{3} + \frac{\varepsilon \cdot {\rm OPT}}{3} \\
								&= (1 + \varepsilon) \cdot {\rm OPT}.
\end{align*}
The number of iterations $T$ required can be further analyzed to:
\begin{align*}
T &\geq \frac{10 \cdot \sigma_r(\widehat{X}_r^\star)}{50} \cdot \frac{3}{\eta \varepsilon {\rm OPT}} \stackrel{(i)}{=} \frac{10 \cdot \sigma_r(\widehat{X}_r^\star)}{50} \cdot \frac{3 \cdot O(L)}{\varepsilon {\rm OPT}} \\
&\stackrel{(ii)}{=} \frac{10 \cdot \sigma_r(\widehat{X}_r^\star)}{50} \cdot \frac{3 \cdot O\left(\tfrac{1}{\tau} + \lambda\right)}{\varepsilon {\rm OPT}} \\
&\stackrel{(iii)}{=} \frac{10 \cdot \sigma_r(\widehat{X}_r^\star)}{50} \cdot \frac{3 \cdot O\left(\tfrac{3mn}{\varepsilon {\rm OPT}} + \tfrac{2 \varepsilon {\rm OPT}}{3 \|X^\star\|_2^2}\right)}{\varepsilon {\rm OPT}} \\ 
&= \frac{10 \cdot \sigma_r(\widehat{X}_r^\star)}{50} \cdot O\left( \tfrac{9mn}{\left( \varepsilon {\rm OPT} \right)^2} + \tfrac{2}{\| X^\star \|_2^2}\right) \\
&= O \left( \sigma_r(\widehat{X}_r^\star) \cdot \left( \tfrac{mn}{\left( \varepsilon {\rm OPT} \right)^2} + \tfrac{1}{\| X^\star \|_2^2}\right) \right)
\end{align*}
where $(i)$ is due to the definition of the step size that $\eta = O\left(\tfrac{1}{L}\right)$,
$(ii)$ is due to the definition $L = \tfrac{1}{\tau} + \lambda$,
$(iii)$ is obtained by substituting $\lambda$ and $\tau$.
\subsection{Proof of Corollary \ref{thm:main2}}

The proof is similar to that of Theorem \ref{thm:main1}.
Using Lemma \ref{lemma:logsumexp2}, we bound $|M - U_TV_T^\top|_\infty$ as follows:
\begin{align*}
|M - U_TV_T^\top|_\infty &\leq \sigma(U_T V_T^\top, \tau) + \tau \log (2 m n)\\
							   &\leq \sigma(U_T V_T^\top, \tau) + \frac{\lambda}{2} |U_T V_T^\top|_2^2 +  \tau \log (2 m n)
\end{align*}
Following similar motions with Theorem \ref{thm:main1}, and setting $\tau = \frac{\varepsilon \cdot {\rm OPT}}{3 \log (2m n)}$, and $T$ and $\lambda$ similar to the $p = 1$ case, we get:
\begin{align*}
|M - U_TV_T^\top|_\infty \leq (1 + \varepsilon) \cdot {\rm OPT}.
\end{align*}
The number of iterations $T$ required follow the same motions as the proof of Theorem 5.1, with a slight difference in the definition of $\tau$.

\section{Connections with other related work}
\citep{collins2002generalization} considers probabilistic extensions of the PCA problem: starting with various generative probabilistic models, one obtains different matrix factorization objectives.
The authors rely on the fundamental work of Csiszar and Tusnady \citep{csisz1984information}, and propose an alternating minimization procedure; 
see also \citep{tipping1999probabilistic, tippingb1999probabilistic}.

\citep{gordon2003generalized2, singh2008unified} show that the differences between many algorithms for matrix factorization can be viewed in terms of a small number of modeling choices. 
Their view unifies methods for Bregman co-clustering, LSI, non-negative matrix factorization, relational learning, to name a few.

While the bilinear factorization $UV^\top$ is common across different problems, there are cases where even a trilinear representation is more preferable, from an interpretation perspective. 
Having constraints over the factors is a another differentiation: An illustrative example of this case is that of matrix co-clustering where we are interested in $M \approx C_1 C_2^\top$, with $C_1$ and $C_2$ being matrices that denote the participation/indicator matrices. 
Our work is quite different to this type of factorizations (\emph{i.e.}, with additional constraints on the factors); we defer the reader to \citep{li2016recovery, gillis2014fast, asteris2016bipartite, veldt2017correlation} for some recent developments on similar subjects.

Finally, there is a recent line of work on robust PCA that further focuses on identifying the (sparse) grossly corrupted elements in $M$; see \citep{xu2010robust, candes2011robust, zhou2011godec, kyrillidis2012matrix, kyrillidis2014matrix, chiang2016robust, gu2016low, yi2016fast}.
That line of work differs from our problem in that, our approach ``models'' the corruption through the penalization of the residual $M - UV^\top$ with an $\ell_1$-norm, while in the aforementioned line of works, one optimizes over the residual $S = M - UV^\top$ in order to minimize the number of ``active" corruptions.
In that sense our model is ``simpler" as we are only interested in identifying the low rank component.

\section{Supportive experimental results}

\begin{table*}[!ht]
\centering
\rowcolors{2}{white}{black!05!white}
\begin{tabular}{c c c c c}
  \toprule
  & & \multicolumn{3}{c}{SVD} \\ 
  & & Time (sec.) & & Error\\
    Rank $r$ & &  \multicolumn{3}{c}{[min, mean, median]} \\  
  \cmidrule{1-1} \cmidrule{3-5} 
 1 & & [2.63e-03, 1.10e-02, 1.08e-02] & & [8.36e-01, 9.02e-01, 9.19e-01] \\ 
 2 & & [3.44e-03, 5.58e-03, 4.25e-03] & & [7.37e-01, 8.60e-01, 8.74e-01] \\ 
 3 & & [4.08e-03, 8.55e-03, 6.67e-03] & & [6.72e-01, 7.51e-01, 7.27e-01] \\ 
 4 & & [2.59e-03, 7.73e-03, 4.47e-03] & & [6.60e-01, 7.31e-01, 7.29e-01] \\ 
 5 & & [2.59e-03, 3.69e-03, 3.63e-03] & & [6.94e-01, 7.21e-01, 7.21e-01] \\ 
 6 & & [2.52e-03, 3.40e-03, 3.11e-03] & & [6.82e-01, 7.22e-01, 7.29e-01] \\ 
 7 & & [2.44e-03, 3.21e-03, 3.29e-03] & & [6.87e-01, 7.35e-01, 7.30e-01] \\ 
 8 & & [2.43e-03, 3.58e-03, 3.32e-03] & & [6.92e-01, 7.36e-01, 7.32e-01] \\ 
 9 & & [2.50e-03, 3.01e-03, 2.97e-03] & & [7.00e-01, 7.27e-01, 7.19e-01] \\ 
 10 & & [1.96e-03, 2.70e-03, 2.84e-03] & & [6.97e-01, 7.61e-01, 7.51e-01] \\ 
  \bottomrule
\end{tabular}

\vspace{0.4cm}
\begin{tabular}{c c c c c}
  \toprule
  & & \multicolumn{3}{c}{\citep{gillis2017low}} \\ 
  & & Time (sec.) & & Error\\
    Rank $r$ & &  \multicolumn{3}{c}{[min, mean, median]} \\  
  \cmidrule{1-1} \cmidrule{3-5} 
 1 & & [6.81e-02, 2.24e-01, 2.28e-01] & & [\textcolor{magenta}{4.91e-01}, \textcolor{magenta}{4.93e-01}, \textcolor{magenta}{4.93e-01}] \\ 
 2 & & [1.55e-02, 2.75e-02, 2.31e-02] & & [5.33e-01, 6.00e-01, 5.96e-01] \\ 
 3 & & [2.42e-02, 5.89e-02, 4.59e-02] & & [5.22e-01, 5.63e-01, 5.44e-01] \\ 
 4 & & [2.69e-02, 4.61e-02, 4.04e-02] & & [5.24e-01, \textcolor{magenta}{5.66e-01}, 5.42e-01] \\ 
 5 & & [4.67e-02, 3.36e-01, 1.48e-01] & & [\textcolor{magenta}{5.04e-01}, 5.36e-01, 5.26e-01] \\ 
 6 & & [6.72e-02, 6.24e-01, 1.34e-01] & & [\textcolor{magenta}{4.98e-01}, 5.20e-01, 5.22e-01] \\ 
 7 & & [5.46e-02, 8.91e-01, 5.47e-01] & & [\textcolor{magenta}{4.90e-01}, \textcolor{magenta}{5.14e-01}, 5.11e-01] \\ 
 8 & & [1.36e-01, 1.66e+00, 5.39e-01] & & [\textcolor{magenta}{4.81e-01}, 5.15e-01, \textcolor{magenta}{5.02e-01}] \\ 
 9 & & [1.90e-01, 2.91e+00, 2.56e+00] & & [\textcolor{magenta}{4.73e-01}, \textcolor{magenta}{4.98e-01}, \textcolor{magenta}{4.89e-01}] \\ 
 10 & & [2.30e-01, 9.60e+00, 4.25e+00] & & [\textcolor{magenta}{4.59e-01}, \textcolor{magenta}{4.97e-01}, 4.79e-01] \\ 
  \bottomrule
\end{tabular}

\vspace{0.4cm}
\begin{tabular}{c c c c c}
  \toprule
  & & \multicolumn{3}{c}{This work} \\ 
  & & Time (sec.) & & Error\\
  Rank $r$ & & \multicolumn{3}{c}{[min, mean, median]} \\  
  \cmidrule{1-1} \cmidrule{3-5} 
 1 & & [2.57e-02, 4.32e+01, 5.44e+01] & & [4.99e-01, 5.82e-01, 5.01e-01] \\ 
 2 & & [2.60e-02, 4.95e+01, 5.44e+01] & & [\textcolor{magenta}{5.04e-01}, \textcolor{magenta}{5.49e-01}, \textcolor{magenta}{5.07e-01}] \\ 
 3 & & [5.20e+01, 5.43e+01, 5.42e+01] & & [\textcolor{magenta}{5.06e-01}, \textcolor{magenta}{5.10e-01}, \textcolor{magenta}{5.10e-01}] \\ 
 4 & & [1.55e-02, 3.67e+01, 5.15e+01] & & [\textcolor{magenta}{5.05e-01}, 5.90e-01, \textcolor{magenta}{5.10e-01}] \\ 
 5 & & [4.17e-02, 7.92e+01, 8.93e+01] & & [5.07e-01, \textcolor{magenta}{5.33e-01}, \textcolor{magenta}{5.13e-01}] \\ 
 6 & & [7.27e+01, 8.03e+01, 7.76e+01] & & [5.02e-01, \textcolor{magenta}{5.08e-01}, \textcolor{magenta}{5.09e-01}] \\ 
 7 & & [1.62e-02, 5.11e+01, 6.52e+01] & & [5.08e-01, 5.84e-01, \textcolor{magenta}{5.08e-01}] \\ 
 8 & & [5.51e+01, 6.55e+01, 6.73e+01] & & [4.95e-01, \textcolor{magenta}{5.09e-01}, \textcolor{magenta}{5.02e-01}] \\ 
 9 & & [5.36e+01, 5.89e+01, 5.77e+01] & & [4.78e-01, 5.06e-01, 5.06e-01] \\ 
 10 & & [1.69e-02, 3.86e+01, 5.23e+01] & & [4.69e-01, 5.94e-01, \textcolor{magenta}{4.75e-01}] \\ 
   \bottomrule
\end{tabular}
\caption{Attained objective function values and execution time. Table includes minimum, mean and median values for 10 Monte Carlo instances.}
\end{table*}

\end{document}